\pdfoutput=1
\documentclass[11pt]{article}

\usepackage{xcolor}

\usepackage{ACL2023}

\usepackage{times}
\usepackage{latexsym}
\usepackage{amsmath}
\usepackage{mathtools}
\usepackage{amsfonts} 
\usepackage{float}
\usepackage{multirow}
\usepackage{graphicx}
\usepackage{arydshln}

\usepackage{enumitem}

\usepackage[T1]{fontenc}
\usepackage[utf8]{inputenc}

\usepackage{microtype}
\usepackage{inconsolata}

\newcommand{\codelocation}[0]{\url{https://github.com/dipta007/SHEM}}

\usepackage{cleveref}
\crefname{section}{\S}{\S}
\crefname{table}{Table}{Tables}
\crefname{figure}{Fig.}{Figs.}
\crefname{algorithm}{Alg.}{}
\crefname{ALC@unique}{Line}{Lines}
\crefname{equation}{Eq.}{Eqs.}
\crefname{appendix}{App.}{Apps.}
\crefformat{section}{\S#2#1#3} 

\usepackage{microtype}

\usepackage{graphicx}
\graphicspath{ {./images/} }

\usepackage{xcolor}
\title{Semantically-informed Hierarchical Event Modeling}

\author{%
\textbf{Shubhashis Roy Dipta, Mehdi Rezaee, Francis Ferraro} \\
  Department of Computer Science and Electrical Engineering\\
  University of Maryland Baltimore County\\
  Baltimore, MD 21250 USA \\
  \texttt{\{sroydip1,rezaee1,ferraro\}@umbc.edu} \\
}

\begin{document}
\maketitle
\begin{abstract}
%%In the domain of event modeling and understanding, we explored hierarchical formation to compress the number of latent variables to model the event script. The core idea of our models was to show that compressing the number of frames can still model the events, as the frames are supposed to be interrelated. We construct a hierarchical, partially observed structured latent model. Our experiments show that our approach does affect the combined perplexity but gives a constant accuracy on inverse narrative cloze, even for the low observed frames.
Prior work has shown that coupling sequential latent variable models with semantic ontological knowledge can improve the representational capabilities of event modeling approaches. In this work, we present a novel, doubly hierarchical, semi-supervised event modeling framework that provides structural hierarchy while also accounting for ontological hierarchy. Our approach consists of multiple layers of structured latent variables, where each successive layer compresses and abstracts the previous layers. We guide this compression through the injection of structured ontological knowledge that is defined at the type level of events: importantly, our model allows for partial injection of semantic knowledge and it does not depend on observing instances at any particular level of the semantic ontology. Across two different datasets and four different evaluation metrics, we demonstrate that our approach is able to out-perform the previous state-of-the-art approaches by up to 8.5\%, demonstrating the benefits of structured and semantic hierarchical knowledge for event modeling. 
\end{abstract}

\section{Introduction}
\label{sec:introduction}
\begin{figure}[t]
    \centering
    \includegraphics{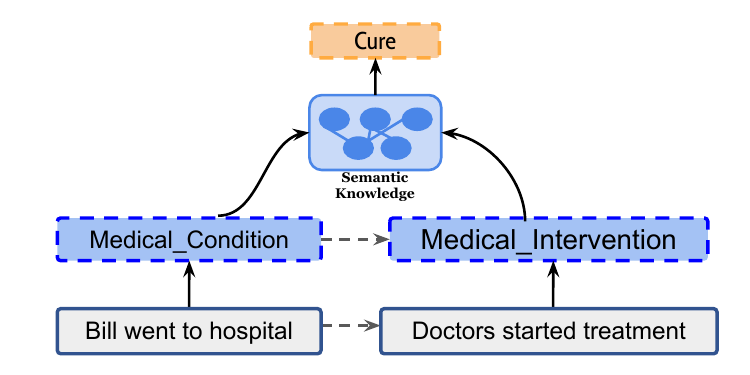}    
\caption{Complex events can be hierarchical. The purple boxes represent the events themselves (as would be reported in a news story). Blue dashed boxes are annotated semantic frames \& the orange dashed box is the more abstract, general frame connecting the ``Medical\textunderscore{}Condition'' and ``Medical\textunderscore{}Intervention'' events. Events and frames are sequentially connected.}
\label{fig:secondary-fig}
\end{figure}

Intuitively, there is a hierarchical nature to complex events: e.g., on \cref{fig:secondary-fig}, there are two events, one involves going to the hospital and another one is getting treatment. Even if important portions may differ, but these two situations have one abstract concept in common: \textbf{Cure} (of a disease). Clearly, there is a connection among the events reported in a situation and they all contribute to a bigger goal (``Cure'' in this case). The main purpose of our work is to exploit this nature of connection to improve event modeling. However, much like linguistic structure, this event structure is generally not directly observed, making it difficult to learn event models that reflect this hierarchical nature.

For high-level inspiration, we look to past approaches in syntactic modeling~\citep{collins-1997-three,klein-manning-2003-accurate,petrov-etal-2006-learning}: we can approach hierarchical event modeling through structured learning, or through richer (semantic) data. %
%Accounting for hierarchical events can be done \textit{structurally} or \textit{semantically}. 
A structural approach accounts for the hierarchy as part of the model itself, such as with hierarchical random variables~\citep{cheung2013pfi,ferraro-2016-upf,weber2018hierarchical,huang2020semi,gao-etal-2022-improving}. %
%On the other hand, 
Richer semantic data provides hierarchical knowledge, such as event inheritance or composition, as part of the data made available to the model and learning algorithm~\citep{botschen-etal-2017-prediction,edwards2022semi,zhang-etal-2020-reasoning}. %

In this work, we provide an approach that addresses both of these notions of hierarchical event modeling jointly. %
Fundamentally, our model is an encoder-decoder based hierarchical model comprised of two layers of semi-supervised latent variable sequence. %
The first layer encodes the events to semantic frames and the next layer compresses down the semantic frames to a more abstract concept. We call these the base and compression layers, respectively.
The base layer operates over the event sequence (the gray boxes in \cref{fig:secondary-fig}); when available, our base layer also considers auxiliary semantic information, such as automatically extracted semantic frames (the blue dashed boxes in \cref{fig:secondary-fig}). %
Meanwhile, the compression layer compresses down the semantic frames to a more abstract concept (orange dashed box in \cref{fig:secondary-fig}) using an existing structued semantic resource (in our paper, FrameNet). %
% We provide targeted semi-supervision through an existing structued semantic resource (FrameNet). %---we do not require additional manual data annotation. %
%, where we leverage a  to provide targeted semi-supervision to those latent variables. %
Our work can be thought of as extending previous work in semi-supervised event modeling~\citep{rezaee-ferraro-2021-event} to account for both structural and semantic hierarchy.

Joining both the structural and semantic approaches together poses a number of challenges. %
First, getting reliable, wide-coverage semantic event annotations can be a challenge. %
Development of semantic annotation resources is time consuming and expensive~\citep{baker1998berkeley,ogorman-2016-red}.\footnote{ %
While prompt-based label semantics~\citep{hsu-etal-2022-degree,huang-etal-2022-unified} are recent successful ways of enabling lower resource learning, these generally are tied to specific tasks and may be limited by what exemplars are given.} %
Part of our solution should leverage existing semantic annotation resources.

Second, although event extraction capabilities have steadily improved, enabling automatically produced annotations to be used directly~\cite{padia2018surface,huang-huang-2021-semantic}, these tools still produce error-laden annotation, especially on out-of-domain text. While rich latent variable methods have been previously developed, adapting them to make use of noisy event extractions can be a challenge. Our learning approach must still be able to handle imperfect extractions. Recent work has shown how neural sequence approaches can do so~\citep{rezaee-ferraro-2021-event}, but there remains a question of how to generalize this. %
Part of our solution should allow for hierarchical semi-supervision. %

\begin{figure*}
    \centering
    \includegraphics{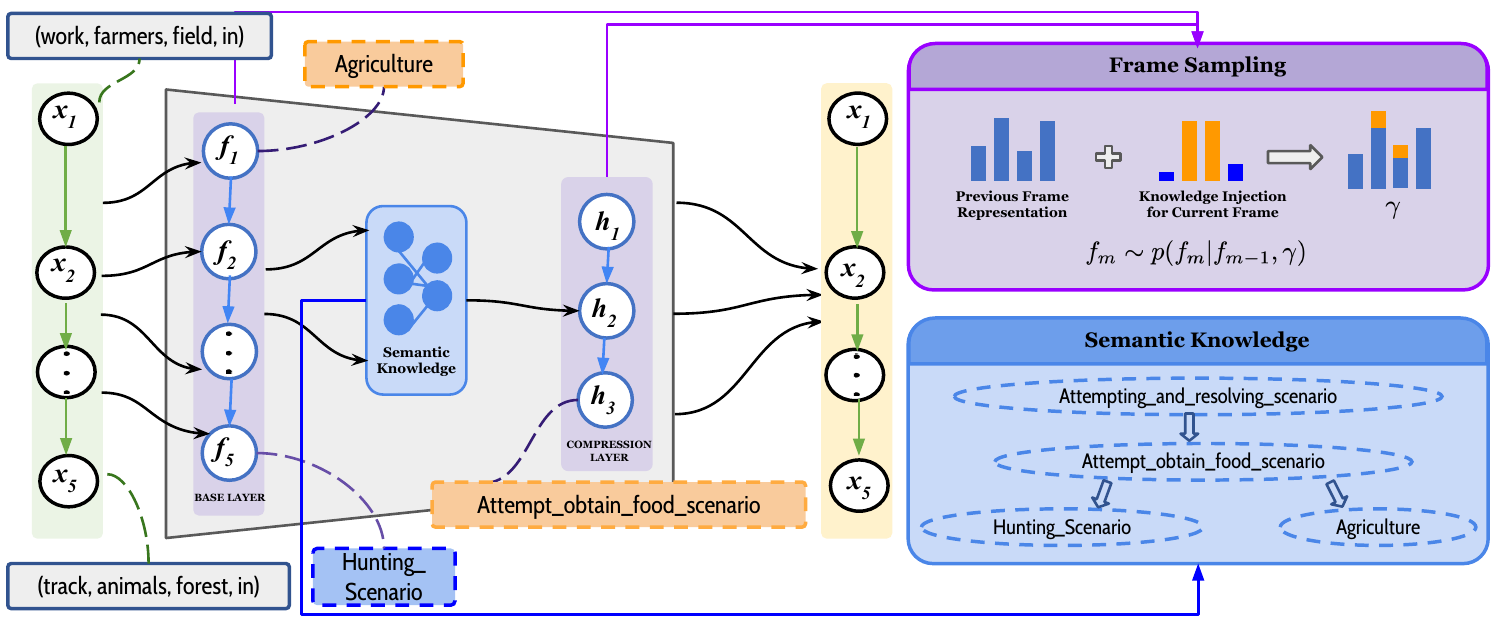}
\caption{ %
An overview of \textbf{S}emantically-informed \textbf{H}ierarchical \textbf{E}vent \textbf{M}odeling (SHEM).
The orange dashed boxes are observed frames \& the blue dashed boxed are masked frames. Top right: a frame is sampled with the injection of observed frames. Bottom right: semantic knowledge graph is shown for 4 nodes with only ``Inheritance'' relations.
\label{fig:main-fig}
}
\end{figure*}

%, including out-of-domain or novel event types, is relatively understudied~\citep{huang2020semi,rezaee-ferraro-2021-event,edwards2022semi}. Especially for handling hierarchical events, the ability to 

%Event extraction from a document is a significant work to summarize documents, find relevant topics, or analyze long documents. Classification with a labeled dataset is a comparatively easy task with the evolution of pre-trained classifiers \citep{kenton2019bert,liu2019roberta}. But finding a labeled dataset is expensive, whereas unlabeled or small labeled is easier to find in the current digital world. But the classification of event frames without any supervision can be a difficult task. The semi-supervision of event frames, which guides the system to detect the unobserved frames, showed a good result in previous research \citep{rezaee-ferraro-2021-event}.

%But even with the extraction of event frames, the number of frames is still the same as the number of event sequences in a process; hence time-consuming and inefficient to understand or analyze the documents. To properly understand, summarize and analyze, we need to cluster semantically similar types of events to their more abstract frame(s). Encoder-Decoder-based attention models have already shown significant results in text classification, language modeling, machine translation, and language generation \cite{vaswani2017attention,devlin2018bert,radford2019language}. 

We present a hierarchical latent variable encoder-decoder approach to address these challenges. %
We ground our work in the FrameNet semantic frame ontology~\citep{baker1998berkeley}, from which we extract possible abstract frames from sequences of inferred (latent) frames. %
This lets us leverage existing semantic resources. %
We develop a semi-supervised, hierarchical method capable of handling noisy event extractions. Our approach %finds important characteristics in an event sequence to encode the representation. %
enables learning how to represent more abstract frame representations. % the number of frames using both partially observed frames and frame relations to identify the higher levels of frames. %
%We have shown that a higher level of frames can express multiple frame knowledge.
Our contributions are:

\begin{itemize}[itemsep=0em,leftmargin=0.5cm]
\item We provide a novel, hierarchical, semi-supervised event learning model.

\item We show how to use an existing rich semantic frame resource (FrameNet) to provide both observable event frames and less observable abstract frames in a neural latent variable model.

%\item We explored a different way of designing compression using the hierarchical model to extract and cluster semantically similar events.

\item Our model can use FrameNet to give a more informed signal by leveraging compression of events when predicting what event comes next, what sequence of events follows an initial event, and missing/unreported events. %

% \item We show that our model can make use of FrameNet frame relations to extract abstract ideas to compress the number of latent variables can effectively cluster similar types of events.\ForDipta{This suggests we're performing clustering experiments -FF}

\item With pre-training only, our model can generate event embeddings that better reflect semantic relatedness than previous works, evincing a zero-shot capability. %which outperforms several similarity tasks.

\item We perform comprehensive ablations to show the importance of different factors of our model.

% \item We show the effect of using a pre-trained language model as encoder and decoder instead of an RNN module.
\end{itemize}
Our code is available at \codelocation{}.

\section{Related Works}
Our work draws on event modeling, latent generative modeling, lexical and semantic knowledge ontologies, and hierarchical modeling.

\subsection{Event Modeling}
% Researchers are trying to understand events from semantic or event sequences.
There have been several efforts to understand events and their relationships with broader semantic notions. %
%Each process consists of multiple sequences, and events in the same process have a connection among them. Therefore, we can even perfectly compress and understand the context by understanding them properly. 
Previous research has explored the use of hierarchical models based on autoencoders for script generation, such as the work of \citet{weber2018hierarchical}. In contrast to their work, instead of a chain-like hierarchy, we have used a multi-layer hierarchy to compress the events to abstract processes. Additionally, our approach allows for semi-supervised training, if such labels are available. Our work has shown that using semi-supervision helps the model to generalize better on both layers.
% \citet{weber2018hierarchical} has used a hierarchical model based on autoencoders to generate the script. This model encodes the event text into a latent variable and regenerates it from those similar to our model. 
% \citet{rezaee-ferraro-2021-event} have used a similar approach to model the event sequences. They have used the gumble softmax technique and injection of partially observed frames to guide the model to generate contextualized event frames. But the number of frames for a process is the same as the number of event sequences in a process, which lacks the capability of compression and generalization. 
In a related study, \citet{rezaee-ferraro-2021-event} used the Gumbel-Softmax technique and partially observed frames to model event sequences and generate contextualized event frames. While their approach is capable of generalizing each event in a sequence, the number of predicted frames in the sequence is equivalent to the number of events. Thus, unlike our approach, it was not designed to compress or generalize the overall event sequence.

% \citet{bisk2019benchmarking} has used event modeling to generate a concrete concept from an abstract concept. They have shown that an abstract concept of cooking can be divided into multiple discrete steps. 
\citet{bisk2019benchmarking} demonstrated the effectiveness of event modeling for generating a concrete concept from an abstract one, using the example of cooking. %
% researchers have used event modeling to predict the types of events of an event process 
Several studies in recent years have utilized event modeling to predict event types
\citep{chen2020you, pepe2022steps, huang2020semi}. 
% In every event process, one particular \textit{action} is done, and one \textit{object} is affected by that action. These papers try to formulate that \textit{action} \& \textit{object} from a event process.
These studies focus on identifying the action and object involved in an event, where the action represents the activity being performed and the object is the entity affected by the action. 

\subsection{Latent Generative Modeling}
Latent generative modeling is a widely-used method for representing data $x$ through the use of high-level, hidden representations $f$. 
% When performing latent modeling, the sequence is encoded at a low level of latent variables that encode high levels of information in a small space. 
% This hidden representation helps to decode the sequence to its original format from the encoded one. 
% Mathematically, if the sequence $x$ is modeled as a hidden representation vector $f$, then the joint probability $p(x, f)$ can be expressed as
Specifically, we express the joint probability $p(x,f)$ as
$
    p(x, f) = p(x|f) p(f).
$
Especially when $f$ is not fully observed, this factorization can productively be thought as a soft grouping or clustering of the data in $x$. %
This equation will serve as the foundation for our approach. %analysis of the relationship between the sequence $x$ and the hidden representation vector $f$.

Maximizing log-likelihood is known to be computationally challenging in this context. %
%To address this issue, \citet{kingma2013auto} proposed the Variational Autoencoder (VAE) framework.
\citet{kingma2014semi} later used a variational autoencoder \citep[VAE]{kingma2013auto} in a semi-supervised manner to learn latent variables, dividing the dataset into observed and unobserved labels. In our case,  instances are partially observed (rather than fully observed or not).
\citet{huang2020semi} used a VAE both to prevent overfitting on seen event types and to enable prediction of novel types. % based on the guidance of seen types.

\subsection{Lexical and Semantic Resources}
\label{sec:semantic-resources}
Multiple resources, such as PropBank~\citep{gildea2002automatic}, OntoNotes~\citep{hovy2006ontonotes}, AMR~\citep{banarescu2013amr}, VerbNet~\citep{schuler2005verbnet}, and FrameNet~\citep{baker1998berkeley}, provide annotations related to event semantics. %
%The vast majority of these resources 
Many consider predicate-argument semantics, such as defining who is performing (or experiencing) an event, and various ways that event may occur. %while extensions are being made (MS-AMR, SemLink),
FrameNet provides detailed predicate-argument characterizations and %\footnote{
%FrameNet, introduced by \citep{baker1998berkeley}, is a linguistic resource that represents the meaning of words in terms of conceptual frames. Each frame in FrameNet consists of three components: (1) the frame itself, which represents a specific conceptual domain or meaning; (2) the frame elements, which are the specific components or attributes that make up the frame; and (3) the lexical units, which are the words or phrases that trigger the activation of the frame in language use.} %
multifaceted relations linking different frames together, such as frame subtyping (e.g., \textit{inheritance}), temporal/causal (e.g., \textit{precedes}, \textit{causative}), and compositionality (e.g., \textit{uses}, \textit{subframe}). %
%These relations allow for the grouping of similar types of events. %
Consider the \textsc{agriculture} frame from \cref{fig:main-fig}: FrameNet defines an \textit{inheritance} relation between it and a \textsc{Attempt\_obtain\_food\_scenario}, which can be thought of as a container grouping together frames all related to a broader scenario of attempting to obtain food, such as \textsc{Hunting\_scenario}. A scenario container frame provides a notion of compositionality, defining potential correlations or alternatives among frames. % %%\footnote{For \textsc{agriculture}, there are additional frame relations \textit{Is perspectived in} (\textsc{Growing\_food}) and \textit{subframe} (\textsc{Food\_gathering} and \textsc{Planting}) that provide additional rich semantic information.}
Due to these rich semantics, we focus on FrameNet in this paper as an exemplar. %

Prior research has shown the utility of FrameNet in predicting the relationship between predicates~\citep{aharon2010generating,ferraro-2017-sem-frames}; frame-directed claim verification~\citep{padia2018surface}; and text summarization~\citep{guan2021frame,han2016text,chowanda2017automatic}. %
Unfortunately, while document-level frames have been of long-standing interest within targeted domains~\cite{sundheim1992muc4,sundheim1996muc6,ebner-etal-2020-multi,du-etal-2021-grit}, development of task agnostic document-level frames has been limited. E.g., while FrameNet defines these compositional-like scenario frames, annotation coverage is limited: In the FrameNet 1.7 data used to train frame parsers, out of nearly 29,000 fulltext annotations, there are only 28 \textbf{annotated} ``scenario'' frames.

\section{Method}
Our core aim is to provide a hierarchical event model that incorporates both structural and semantic hierarchy. We call our model SHEM (\textbf{S}emantically-informed \textbf{H}ierarchical \textbf{E}vent \textbf{M}odeling). An overview is in \cref{fig:main-fig}, where an observed event sequence (green $x_i$) is latently modeled as multiple sequences of semantic frames ($f_i$ and $h_j$), augmented by a semantic resource. % along helps to cluster\todo{check ``clustering''} the events and represent abstract ideas. 

We examine the strengths and limitations of structural and semantic hierarchy. Our experiments explore the effect of compressing the number of frames on ability to predict what happens next in an event sequence, and, given an initial seed event, how an event sequence is likely to unfold.  We also extend our work to show how our model can produce better intrinsic event representations.

%In our work, the encoder encodes the main input text into a number of frames, i.e., 5, and then again on the compression layer encodes both the input text and base layer latent variable embeddings to fewer frames, i.e., 3. The decoder reconstructs the input text from the predicted frames for both layers. The loss is calculated on both layers using the reconstructed text and for predicted frames.

\subsection{Model Setup}
Our model is a sequence-to-sequence hierarchical model (§\ref{hierarchical}). It is comprised of two layers (a base and a compression layer) of an encoder \& decoder (§\ref{sec:input-output}). %---processes an input event sequence to encode the events into latent variables and regenerate the events from the latent variables. %The base layer learns how to encode the event sequence as a corresponding sequence of semantic frames, while the compression layer encodes the sequences into fewer latent variables. 
During training (§\ref{training}), we provide the model partially observed semantic frames in the base layer in order to guide it in encoding event sequences into latent variables. In the compression layer, we use ontologically-defined frame relations to extract semantically similar frames from the predicted frame of the first layer. These semantically similar frames guide the compression layer of the model to infer appropriate abstract frames. %By using related frames, the model can get a sense of the higher-level frames that can be cluster of multiple frames, which in terms will compress the number of frames.\todo{check ``clustering''}

\subsection{Input and Output}
\label{sec:input-output}
The input to our model is event sequences. %
Each sequence is defined by $M$ event tuples $(x_1, x_2, ... x_M)$. %
For comparability \citep{weber2018hierarchical,rezaee-ferraro-2021-event}, we represented each event as a tuple $x_m$ of four lexical words: a predicate, a subject, an object, and an optional event modifier. We assume an event tuple can be associated with a more general semantic frame. %We refer to the frame for $x_m$ via $f_m$. 
For example, in \cref{fig:main-fig}, the first event (``work farmers in field'') can be linked to the FrameNet \textsc{Agriculture} frame. We assume that each event \textit{can} be linked but do not require this. %
Some frames might be masked, subject to a fixable observation probability. This allows us to test how our model behaves when semantic data may be missing or incorrect (due to, e.g., an extraction error); in \cref{fig:main-fig}, this can be seen for the event ``track animals in forest'' event, where a potential corresponding frame---``Hunting\_Scenario''---is masked. %
This results in a corresponding sequence of (partially) observed frames $(f^*_1, f^*_2, ... f^*_M)$. % 
The base layer uses these event tuples ($x_i$) to softly predict the frames ($f_i$) and then reconstruct the input sequence based upon those inferences. To capture additional semantic knowledge, both in traning and testing, we query FrameNet to extract more abstract frames ($h_i$) for the predicted frames from the base layer, such as ``Attempt\_obtain\_food\_scenario.'' The compression layer uses that abstract frame $h_i$ with the original event frames $f_i$ to softly group the events; for additional training signal, the compression layer is also trained to reconstruct the original event sequence.

\vspace{.4em}
%\subsection{Encoder} \label{encoder}
\noindent\textbf{Encoder}
%Each layer must be able to properly encode its input. %
The base layer embeds each token in the input event sequence, while, by default, the compression layer embeds each predicted frame from the base layer. An attention module is used to find the important parts of event sequences during prediction of frames.
As our experiments validate, the encoder can be flexible, e.g., a bi-GRU or a Transformer-based large language model. 
%From the input of event sequences and the corresponding frames, the encoder first generates embeddings from the event sequences, and a Gated Recurrent Unit (GRU) \citep{cho2014properties} extracts the hidden feature from the event sequences. 
%We view the per-token embeddings as newly computed hidden features, which are then passed to the linear latent nodes to extract the latent variables. 
%Next, we discuss each layer's encoder and how we inject semantic knowledge into the modeling process. 

%\subsection{Decoder} \label{decoder}
\vspace{.4em}
\noindent\textbf{Decoder}
%In our model, the decoder part is comparatively easier than other modules, and it is the same on both of the layers. For this model, we have used an
This is a standard auto-regressive model that generates tokens of an event sequence from left to right. Unless otherwise specified, the predicted frame embeddings are given as input to the decoder. %Like any auto-regressive model, previously generated decoder output and previous input texts are given as input to the decoder. %An attention module is used to find the important words from the given latent embeddings predicted by encoder. %
%Each layer tries to reconstruct the input text, and loss was generated individually for each layer, which then accumulated and back-propagated through the whole model, updating the model parameters.
See \cref{sec:more-model-details} for additional details.

\subsection{Hierarchical Model} \label{hierarchical}

We use two layers of an encoder-decoder: (i) a base layer ($f_i$s in \cref{fig:main-fig}) and (ii) a compression layer ($h_j$s in \cref{fig:main-fig}). The base layer is responsible for encoding the input event sequence into a sequence of semantic frames, while the compression layer is responsible for re-encoding the base layer's semantic frames into more abstract representations. In \cref{fig:main-fig}, the base layer must infer ``Agriculture'' \& ``Hunting\_Scenario'' from the input and observed frames; the compression layer must associate those frames with ``Attempt\_obtain\_food\_scenario.'' Our model is extendable to an arbitrary number of compression layers. Experiments with multiple compression layers showed that a single compression layer was sufficient for strong performance. 

%Reconstruction solely depends on the frame samples generated by the encoder. %In keeping with the previous work \citep{rezaee-ferraro-2021-event}, 
Given our encoder-decoder setup, inferring frame values means sampling a discrete random variable within a neural network. This must be done at both the base and compression layers. To do so, we sample frames from an ancestral Gumbel-Softmax distribution~\citep{jang2016categorical,rezaee-ferraro-2021-event}: each sampled frame $f_i$ depends on the previously sampled frame $f_{i-1}$ and an attention weighted embedding of that layer's encoder representation. Due to space, we refer the reader to \citet{rezaee-ferraro-2021-event}. %
%The encoder learns a good representation \(\gamma_i\) for each event, and samples the current frame, $f_m$ from the Gumbel-Softmax distribution parameterized by \(\gamma_m\).

\vspace{.4em}
\noindent\textbf{Base Layer} %
The base layer encodes the event sequences in the same number of latent variables with the guidance of the observed frames. %
On the base layer, partially observed frames are fed to the model. These frames depend on the observation probability; e.g., 40\% observed frames mean that 60\% of the event frames will be masked, and the remaining 40\% would be observable by the model as guidance. %
This masking, which we formalize as part of our experiments, reflects the fact that we may not always have access to sufficient semantic knowledge. 
To guide the base layer, a one-hot encoding of the observed frames is ``injected'' (added to the Gumbel-Softmax parameters), as done by \citet{rezaee-ferraro-2021-event}. %
The number of frames is the same as the number of event sequences, so one frame for each node is passed. 

\vspace{.4em}
\noindent\textbf{Compression Layer} %
\citet{rezaee-ferraro-2021-event} showed that providing some frame injection guidance helps learning. %
The compression layer aims to provide guidance to the modeling through fewer, more abstract semantic frames. %
However, while this is possible for the base layer, where we assume every event tuple \textit{could} have a frame, we do not assume this for the compression layer. %that a compressed, potentially more abstract, sequence of frames is observed. %
This in part is reflective of the lack of annotated training samples for some of these more abstract frames (see \cref{sec:semantic-resources}), limited beyond-sentence frame extraction tools, and our own motivation to not require beyond-sentence annotation or extraction tools.

To provide guidance, but prevent reliance on potentially missing auxiliary semantic knowledge, we extract the \textit{inferred} frames from the base layer with the external frame ontology (rather than whatever frames may have been provided to the model). %
For each inferred frame $f_i$, we extract possible abstract frames using the FrameNet relations defined for it. E.g., since there is a frame relation between \textsc{agriculture} and \textsc{Attempt\_obtain\_food\_scenario}, if $f_i$ is \textsc{agriculture}, \textsc{Attempt\_obtain\_food\_scenario} may be an abstract frame. %
In the case of multiple abstract frames, one single frame is chosen randomly. A special frame token (not in FrameNet) is passed if no related frames can be extracted. %\footnote{
%\footnote{For the compression layer frames, we have done a number of experiments to explore which relation or signal of input has more effect on the learning. The results of those models are shown in \cref{result}.} 
%
%The frame itself and the abstract frame are extracted for each predicted frame. %
% In total, $2M$ frame values are fed to the compression layer. %
% We use a multi-hot encoding of the selected frames for injection in inferring the compression layer's frame values $h_j$. %
% } %
%But on the compression layer, each node has a $2 * M$ number of frames, so multi-hot encoding is done where each row can have at most $2*M$ number of 1s and others as $0$. 
Each compression node $h_j$ has an attention module, attending over the base layer's inferred frames $f_1, \ldots, f_M$, helping capture ontological hierarchy. %the event sequences and passed frames. 

While the compression layer can serve as an event model in its own right (due its own decoder), its primary purposes are to help \textbf{capture the ontological hierarchy and provide feedback to the base layer}. It does this directly (predict the extracted abstract frames, given the base layer's inferred frames as input), and via its decoder.

\vspace{.4em}
\noindent\textbf{Guidance for Abstract Frames}
To guide the compression layer to learn more abstract frames and help the base layer generalize, we injected the FrameNet-defined parents of the frames predicted from the base layer. E.g., if the base layer prediction is ``Temporary\_Stay'' and a related frame is ``Visiting,'' we inject both to the compression layer. %Due to the multiple numbers of frames for each node, we have injected a multi-hot vector instead of a one-hot vector like the base layer. 
In contrast to existing work relying on single samples, %we conjectured that additional samples would provide stronger signal during learning; 
early experiments showed that averaging two Gumbel-Softmax samples yielded better results.

\subsection{Training} \label{training}
{{During training, input is passed to the base layer with partially observed frames depending on the observation probability. The first layer encoder encodes the input sequence with the guidance of the partially observed frames to generate a latent variable representation $(f_{i})$. This predicted latent variable $(f_i)$ is then passed through the decoder to regenerate text. The predicted frames from the first layer and their parent frames are passed to the second layer encoder; it then encodes it to fewer numbers of latent variables, $(h_j)$ which is used in the decoder. Loss is computed at both layers.}}
%, (i) the reconstruction loss $(\mathcal{L}_\beta)$, (ii) the KL-divergence loss $(\mathcal{L}_\alpha)$, and (iii) the frame classification loss $(\mathcal{L}_\gamma)$. Frame classification loss is calculated only for the base layer. 
% In our model, the loss is formulated as a weighted combination of three types of losses: a reconstruction loss $\mathcal{L}_{\beta_j}$, which generates $x$ according to inferred latent variables from layer $j$; the KL divergence loss $\mathcal{L}_{\alpha_j}$, which computes the KL divergence between the prior and variational distributions of each layer $j$; and a frame classification loss $\mathcal{L}_{\gamma}$, which uses the base layer's variational distributions to compute a cross entropy loss for the frames that were observed. %
% The weights are minimally tuned on dev data.\ForDipta{Is this correct?} %
% See \cref{sec:loss-formulation} for a full mathematical formulation of our loss. %

We employ a linear combination of three different loss functions: the reconstruction loss, the KL divergence loss, and a frame classification loss. The reconstruction loss %denoted as $\mathcal{L}_{r_j}$, 
is used to generate the input event sequence based on the inferred latent variables from each layer. The KL divergence loss %, represented as $\mathcal{L}_{\text{KL}_j}$, 
calculates the KL divergence between the prior and variational distributions for each layer. Finally, the frame classification loss %denoted as $\mathcal{L}_{c}$, 
guides the base layer to accurately classify the observed frames.
See \cref{sec:loss-formulation} for a full formulation of our loss. %

%The final loss $(\mathcal{L})$ is formulated as follows.

\section{Experimental Setup}
\label{sec:experimental_setup}
% In this section, we detail the experimental setup and 
% results for our proposed model. 
We describe the dataset, then baselines (\cref{im_base}), we used for our core experiments. %
We explored the effectiveness of latent parent frames (\cref{exp_3.5}) and %lexical signal (\cref{exp_3.12}), 
frame relations (\cref{exp_3.10}). %grouping of frame properties (\cref{exp_3.13}) 
We show how our model accounts for missing events (\cref{exp_masked}). To further show the effectiveness of our model, we show how to extend our approach to provide effective representations for event similarity tasks (\cref{sec:sentence-embeddings}). %
We provide supplementary results and experiments in the appendix.

\vspace{.3em}
\noindent\textbf{Dataset}
%\subsection{Dataset} \label{dataset}
We used a part of the Concretely Annotated Wikipedia dataset \citep{ferraro2014concretely}, which is a version of English Wikipedia that provides automatically produced FrameNet semantic frame parses~\nocite{das2014frame} to enable easier subsequent examination of semantic frames. This has existing splits of training (457k), validation (16k), and test (21k) event sequences, where each training sequence has at least one extracted frame. For comparability with past approaches, we truncated documents to the first 5 events. %Sequences with more events are trimmed off after five events. 
We used a vocabulary size of 40k for event sequences (predicates and arguments) and the 500 most common semantic frames, which is consistent with prior work and has more than 99\% coverage of automatically extracted frame types. 

\subsection{Implementation and Baselines} \label{im_base}
We use five latent variables in the base layer and three in the compression layer; these values were determined in early dev experiments. We represent the probability of observing an event's frame on the base layer with an observation probability $\epsilon$. With $\epsilon$ likelihood, an event's frame will be observed, and with $(1 - \epsilon)$ probability, an event's frame will be masked. This is meant to emulate how sufficiently accurate, extractable semantic knowledge may not always be available. This $\epsilon$ was fixed prior to training each model. Frames are \textbf{only observed during training, and never during evaluation}. More implementation details, including specific hyperparameter values and architectural decisions, are in \cref{sec:implementation-details}. %
We present extensive ablation experiments in \cref{sec:ablation}. These experiments provide further insight into our modeling decisions.

\paragraph{Baselines}
Most of our experiments (\cref{exp_3.5,exp_3.10,exp_masked}) compare our model with the existing methods:
%\begin{itemize}
%\item 
First, \textbf{HAQAE} \citep{weber2018hierarchical}, which employs a single layer, chain-based method for hierarchical modeling. It is designed purely as an unsupervised approach, and so we cannot provide frame guidance to it. We retrained this model on our event sequences. 
%
%\item 
Second, \textbf{SSDVAE} \citep{rezaee-ferraro-2021-event}: this is most similar to ours and effectively just the base layer. %The authors have used the same encoder-decoder with the same number of hidden states with the same pre-trained embedding.  
%\end{itemize}
For fairness, we use the same hidden state size and pre-trained embeddings across our models and baselines.

\section{Result and Discussion} \label{result}
We compute standard event modeling metrics: perplexity, to measure how well the model can predict the \textit{next} event, and inverse narrative cloze (INC) score~\citep{weber2018hierarchical}. % for Wikipedia test data. 
In INC, a single seed event is given, and the model must select what the next five events are to follow it. The model is given six choices (giving random performance accuracy of $16.7\%$). Both have been used by our baselines and allow us to assess the effectiveness of our model. We average results over four runs with different seeds, unless otherwise specified.

% For all of the experiments, we have used FrameNet to extract event relation semantics. Though there were many other lexical databases available \citep{propbank, miller1995wordnet}, the main shortcomings on most of them is the absence of relation semantics. 
% We reported perplexity with the test dataset to measure the model performance. Additionally, we have reported the inverse narrative cloze score for Wikipedia test data.

% exp 3.5
\subsection{Is Frame Inheritance Sufficient?} \label{exp_3.5}

\begin{table}[t]
\small
%\resizebox{.98\columnwidth}{!}{%
\begin{tabular}{c|c|cc}
\hline
Model  & $\epsilon$   & Perplexity ($\downarrow$) & INC Score ($\uparrow$) \\ \hline
HAQAE                   & -                    & 21.38 $\pm$ 0.25             &  24.88 $\pm$ 1.35  \\ \hdashline
SSDVAE                  & \multirow{3}{*}{0.9} & 19.84 $\pm$ 0.52              &  35.56 $\pm$ 1.70  \\ 
ours: inf. frame &                      & \textit{\bf 19.39 $\pm$ 0.3}    & 41.35 $\pm$ 4.25   \\
% ours: lexical &                      & \textit{\textbf{19.12 $\pm$ 0.53}}    & \textit{41.35 $\pm$ 3.19}   \\ 
\hdashline

SSDVAE                  & \multirow{3}{*}{0.7} & 21.19 $\pm$ 0.76 & \textit{39.08 $\pm$ 1.55}   \\ 
ours: inf. frame                 &                      & \textit{20.26 $\pm$ 1.36}    & 35.86 $\pm$ 3.43 \\ 
% ours: lexical                   &                      & \textit{21.52 $\pm$ 1.48}    & 35.61 $\pm$ 4.72   \\ 
\hdashline

SSDVAE                  & \multirow{3}{*}{0.5} & 31.11 $\pm$ 0.85 & \textit{40.18 $\pm$ 0.90}   \\ 
ours: inf. frame                 &                      & \textit{22.16 $\pm$ 1.62}    & 37.3 $\pm$ 3.33 \\
% ours: lexical                   &                      & \textit{25.02 $\pm$ 1.31}    & 37.8 $\pm$ 3   \\ 
\hdashline

SSDVAE                  & \multirow{3}{*}{0.4} & 33.12 $\pm$ 0.54 & \textit{47.88 $\pm$ 3.59}   \\ 
ours: inf. frame                 &                      & \textit{24.02 $\pm$ 1.28}    & 43.25 $\pm$ 4.97 \\
% ours: lexical                   &                      & \textit{27.06 $\pm$ 0.94}    & 39.2 $\pm$ 1.23 \\ 
\hdashline

SSDVAE                  & \multirow{3}{*}{0.2} & 33.31 $\pm$ 0.63 & 44.38 $\pm$ 2.10   \\ 
ours: inf. frame                 &                      & \textit{30.15 $\pm$ 2.73}    & \textit{\bf 49.53 $\pm$ 1.56} \\
% ours: lexical                   &                      & \textit{33.6 $\pm$ 1.84}    & 46.53 $\pm$ 2.84 \\ 
\hline
\end{tabular}%
%}
\caption{Perplexity (lower is better) and Wikipedia Inverse Narrative Cloze Score (higher is better) for test data. Per observation probability ($\epsilon$), the best is in \textit{italic} form. The best overall is \textbf{bold} form. See \cref{exp_3.5}.} %, \cref{exp_3.12}}
\label{exp_3.5_12_combined}
\end{table}

We first investigate whether frame inheritance is sufficient for learning our hierarchical model. %
We report the inferred frame variant previously described: the base layer first infers the latent frames; then we extract the parents of those inferred frames; and we then inject both these parent frames and base layer predicted frames in the compression layer. %
The compression layer is dependent on the inferred frames, rather than lexical signal. %
Results are in \cref{exp_3.5_12_combined} (supplemental results in \cref{exp_3.5_12_ppl,exp_3.5_12_winv} in the appendix). %
% Where the variants differ is in the input to the compression layer. %
% In the inferred frame variant, the input to the compression layer is the inferred base layer frames. %The aim was to determine whether the incorporation of parent frame knowledge would enhance the performance of the compression layer and improve the base layer generation capability.
% In the lexical variant, it is the lexical embedding of the original event sequence. %
% These variants examine how the inferred frames impacts model performance.
We also experimented with a lexical variant, where the input to the compression layer is an embedding of the original input event tuple rather than the inferred frames. Due to space constraints, these detailed comparisons are in \cref{exp_3.5-app}. The compression layer alone has suboptimal performance on both lexical and inferred frame models, but the signal from compression layer helped the base layer to achieve better performance. Both SSDVAE and HAQAE (no compression layer) did worse for all observation probabilities. This shows the inferred frames and semantic relations from the base layer are important for hierarchical modeling.\footnote{In particular, \cref{fig:improvement-from-compression} in the appendix shows how the compression layer can demonstrate its own generative capabilities, in addition to providing supervisory signal to the base layer.}

%The perplexity result of this experiment is in \cref{exp_3.5_12_combined}, with additional per-layer results in \cref{exp_3.5_12_ppl}. 
Our model's base layer perplexity consistently outperformed the other models. %
Additionally, we see that our approach is better able to handle lower supervision than SSDVAE: as the observation probability decreases (fewer observed semantic frames), perplexity increases drastically for SSDVAE. In contrast, if we look at the ``ours: inf. frames'' perplexity, we see that any performance degradation in our model is less severe, and that in all cases our approach still outperforms the previous SOTA results. This shows the effectiveness of the compression layer in guiding the base layer reconstruction, even with limited semantic observation.

%INC is summarized in  \cref{exp_3.5_12_combined}, with per-layer scores reported in \cref{exp_3.5_12_winv}. %
Looking at INC, with either a lot ($\epsilon=0.9$) or a little ($\epsilon=0.2$) of semantic observations, our approach outperforms the existing approaches, demonstrating the ability to model longer event sequences. %
The best overall INC performance occurs with our hierarchical model with a low amount of supervision. %
This is a good result, as it suggests our model can make use of limited semantic extractions and still provide effective long-range modeling. %
When some, but not necessarily most, of the frames may be observed, the non-hierarchical SSDVAE approach provides strong performance. %
This suggests that while frame inheritance (e.g., \textit{IS-A} type relations) can be helpful for certain elements of hierarchical event modeling, it is not sufficient. %
However, as we will see in the next section, more considered use of semantic relations defined in FrameNet can drastically boost our model's performance, surpassing SSDVAE.
\subsection{Relations Beyond Inheritance} \label{exp_3.10}
We have shown that inheritance relations are helpful but not sufficient. As FrameNet reflects other relations, like causation, (temporal) ordering, and multiple forms of containment/composition, we explore whether six different frame relations significantly affect the predictive abilities of our model. %

We also consider two special cases: first, whether different types of relations are complementary by grouping these select relations.\footnote{ %
We aggregate frames connected via the Inheritance, Using, Precedes, Causative\_of, Inchoative\_of, and Subframe relations. %
We selected these given their direct connections to well-studied relationships across event semantics. %
} %
We refer to this as \textit{grouping} in \cref{exp_3.10_combined}. %
Second, whether the compositional ``scenario'' frames in FrameNet provide a strong signal (\textit{scenario-only} in \cref{exp_3.10_combined}). %
In FrameNet, frames that introduce a broader, abstract concept rather than an isolated one can be labeled as a ``scenario'' frame: e.g., \textit{COMMERCE\_SCENARIO} consists of buying, selling, business, having an agreement, and so on. %
For this, we only extracted an abstract frame for the compression layer if it was labeled as a ``scenario.'' 
%As for the compression layer, our main target was to compress the data to a more abstract concept; for this experiment, we tried to see if the scenario frames help generalize the latent frames and help the regeneration or event modeling.

\begin{table}[t]
\resizebox{\columnwidth}{!}{%
\begin{tabular}{c|c|c|cc}
\hline
\multirow{2}{*}{Model} & \multirow{2}{*}{Frame Relation} & \multirow{2}{*}{$\epsilon$} &  Next Event  & Event Sequence Pred.  \\
 &  & & Pred. (Perplexity)   & (Wiki INC Accuracy) \\ \hline
HAQAE                  & - & -                    & 21.38 $\pm$ 0.25        & 24.88 $\pm$ 1.35 \\ \hline

SSDVAE                  & - & 0.9     & 19.84 $\pm$ 0.52    & 35.56 $\pm$ 1.70 \\ \hdashline
\multirow{9}{*}{ours}   & Inheritance &    \multirow{8}{*}{0.9}             & 19.39 $\pm$ 0.53    & 41.35 $\pm$ 4.25   \\ 
                        & Using &                       & 19.39 $\pm$ 0.51    & \textit{43.23 $\pm$ 2.51}   \\ 
                        & Precedes &                    & 19.57 $\pm$ 0.58    & 41.43 $\pm$ 3.02   \\
                        %& Metaphor &                    & 19.62 $\pm$ 0.75    & 41.92 $\pm$ 3.93   \\
                        %& See\_also &                   & 19.55 $\pm$ 0.72    & 42.67 $\pm$ 1.49   \\
                        & Causative\_of &               & 19.42 $\pm$ 0.57    & 41.38 $\pm$ 2.23   \\
                        & Inchoative\_of &              & 19.28 $\pm$ 0.32    & 41.35 $\pm$ 3.47   \\
                        & Perspective\_on &             & 19.76 $\pm$ 0.97    & 40.53 $\pm$ 2.04   \\
                        & Subframe &                    & 18.91 $\pm$ 0.15    & 40.35 $\pm$ 2.91   \\
                        %& ReFraming\_Mapping &          & 19.56 $\pm$ 0.94    & 43.8 $\pm$ 4.02   \\
                        & \textit{grouping} &          & 19.44 $\pm$ 0.5    & 40.76 $\pm$ 2.86   \\
                        & \textit{scenario-only} &          & \textbf{\textit{18.81 $\pm$ 0.5}}    & 42.29 $\pm$ 2.86   \\ \hline

SSDVAE                 & - & 0.2 & 33.31 $\pm$ 0.63         & 44.38 $\pm$ 2.10 \\ \hdashline
\multirow{9}{*}{ours}   & Inheritance &   \multirow{8}{*}{0.2}     & \textit{30.15 $\pm$ 2.73}    & 49.53 $\pm$ 1.56  \\ 
                        & Using &                       & 31.37 $\pm$ 2.08    & 49.72 $\pm$ 1.73   \\
                        & Precedes &                    & 32.62 $\pm$ 1.65    & 47.92 $\pm$ 2.25   \\
                        %& Metaphor &                    & 32.92 $\pm$ 2.08    & 47.25 $\pm$ 3.81   \\
                        %& See\_also &                   & 31.83 $\pm$ 2.78    & 47.77 $\pm$ 3.61   \\
                        & Causative\_of &               & 31.82 $\pm$ 3    & \textbf{\textit{49.85 $\pm$ 0.84}}   \\
                        & Inchoative\_of &              & 32.65 $\pm$ 1.4    & 48.03 $\pm$ 3.35   \\
                        & Perspective\_on &             & 33.2 $\pm$ 1.47    & 47.85 $\pm$ 3.53   \\
                        & Subframe &                    & 32.78 $\pm$ 2.09    & 47.88 $\pm$ 3.31   \\
                        %& ReFraming\_Mapping &          & 31.34 $\pm$ 2.76    & 49.05 $\pm$ 1.54   \\
                        & \textit{grouping} &          & 28.17 $\pm$ 2.26    & 48.88 $\pm$ 1.37   \\ 
                        & \textit{scenario-only} &          & 32.01 $\pm$ 0.7    & 48.1 $\pm$ 2.22   \\ \hline
\end{tabular}%
}
\caption{Using frame relations beyond inheritance for the compression layer can lead to drastic improvements in both perplexity (lower is better) and Wikipedia Inverse Narrative Cloze Score (higher is better).
See \cref{exp_3.10}. For detailed result with all the layers, please refer to appendix (\cref{sec:frame-relations-appendix,sec:exp_3.19_appendix,sec:exp_3.13-app}). %
}
\label{exp_3.10_combined}
\end{table}

We trained separate models (with three random seeds) for each frame relation to explore the effect of individual frame relations on the result. %
We focus on higher ($\epsilon=0.9$) and lower ($\epsilon=0.2$) frame observation cases. %
\Cref{exp_3.10_combined} shows our main results, with detailed results in the appendix (\cref{sec:frame-relations-appendix,sec:exp_3.19_appendix,sec:exp_3.13-app}). %Perplexity is summarized in \cref{exp_3.10_ppl}. 
Lower observation ($\epsilon=0.2$) is consistently better than the previous state-of-the-art on the base and overall versions. For $\epsilon=0.9$, base layer performance is generally improved. %
This reaffirms our previous results that even with limited semantic guidance, the compression layer provides valuable feedback to the base layer. 

%Wikipedia Inverse Narrative cloze is in \cref{exp_3.10_winv}. Our model achieves significantly better results than the previous state-of-the-art in both the base and total cases. This demonstrates the generalization ability of our model to understand the connections between events and infer abstract meanings.

The results for the two special relations in \cref{exp_3.10_combined} (\textit{grouping} and \textit{scenario-only}) are consistent with our previous results—our approach outperforms the state-of-the-art result. %, whereas the compression layer helps the base and overall model to make a more informed choice. %
Neither grouping nor the scenario-only variant provides large additional benefit beyond the individual frames in that group. %
%While our model---with any of the examined---outperforms the baseline, the differences across frame relations are small. %
Given this and the small variation in base layer performance depending on what frame relations we use, these results suggest that the existence of broader assocations that these relations enable are very helpful. % may richly defined relations, beyond simple inheritance, is helpful. %; in the next section, we examine in greater detail which types of frame relations are most helpful. %
This would suggest that semantically-aware event modeling could benefit from broader semantic resource coverage, with future work examining how best to encode the semantics of \textit{any particular} relation. % may not matter as much to event modeling as 

\begin{table}[t]
\centering
\resizebox{\columnwidth}{!}{%
\begin{tabular}{c|c|ccc}
\hline
\multirow{2}{*}{Model}  & \multirow{2}{*}{$\epsilon$}   &  \multicolumn{3}{c}{Perplexity (Masked Test Data)}\\ 
                        &                               & Base Alone      & Compression Alone       & Base+Compr. \\ \hline

\textit{SSDVAE}         &       \multirow{3}{*}{0.9}    & 152.44 $\pm$ 3.45                 &        -               & -\\
\textit{grp}            &                               & \textit{61.1 $\pm$ 1.83}     & 94.76 $\pm$ 1.96             & 76.08 $\pm$ 0.76 \\
\textit{scn}            &                               & 63.48 $\pm$ 4.43     & 80.94 $\pm$ 7.44             & 71.6 $\pm$ 4.12 \\ \hdashline

\textit{SSDVAE}         &       \multirow{3}{*}{0.7}    & 163.08 $\pm$ 4.52                 &        -               & -\\
\textit{grp}            &                             &  63.5 $\pm$ 3.49     & 86.23 $\pm$ 0.7             & 73.98 $\pm$ 2.04 \\
\textit{scn}            &                                & \textbf{\textit{60.06 $\pm$ 1.68}}     & 78.36 $\pm$ 4.52             & 68.58 $\pm$ 2.3 \\ \hdashline

\textit{SSDVAE}         &       \multirow{3}{*}{0.5}    & 182.63 $\pm$ 6.11                 &        -               & -\\
\textit{grp}            &                                               & 79.74 $\pm$ 1.79     & 83.81 $\pm$ 0.96             & 81.75 $\pm$ 1.13 \\
\textit{scn}            &                               & \textit{76.01 $\pm$ 5.56}     & 78.7 $\pm$ 1.63             & 77.33 $\pm$ 3.65 \\ \hdashline

\textit{SSDVAE}         &       \multirow{3}{*}{0.4}    & 201.55 $\pm$ 4.1                 &        -               & -\\
\textit{grp}            &                               &  84.17 $\pm$ 4.45     & 81.49 $\pm$ 0.14             & 82.8 $\pm$ 2.13 \\
\textit{scn}            &                                 & \textit{73.77 $\pm$ 7.87}     & 80 $\pm$ 1.89             & 76.77 $\pm$ 4.89 \\ \hdashline

\textit{SSDVAE}         &       \multirow{3}{*}{0.2}    & 212.93 $\pm$ 2.54                 &        -               & -\\
\textit{grp}            &                               & 89.73 $\pm$ 4.67     & 77.32 $\pm$ 0.72             & 83.28 $\pm$ 2.38 \\
\textit{scn}            &                               & 83.86 $\pm$ 2.74     & 81.2 $\pm$ 1.17             & \textit{82.52 $\pm$ 1.93} \\ \hline
\end{tabular}%
}
\caption{Perplexity (lower is better) for the grouped and scenario-based models in the scenario-masked evaluation. For each $\epsilon$, the best score is \textit{italicized}. Best overall is \textbf{bold}. These results indicate how our approach can make use of related frames to better model sequences involving missing events. See \cref{exp_masked}. 
}
\label{exp_masked_ppl}
\end{table}

\subsection{Predicting Missing Events} \label{exp_masked}
Previously, we have looked at how using the observation probability can help us mask frames and semi-supervised learning. In this experiment, 
we examine the robustness of our model with respect to missing events in an input sequence along with the frame masking depending on observation probability. We first identify sequences (in our training, dev, and test data) where two events have different frames $f_i$ and $f_j$ that are contained within the same scenario frame. We train normally, but to evaluate, we remove an event $e_j$ associated with a scenario-connected frame $f_j$ from the input. Given this impoverished input, we require the model to generate the full, unmodified sequence. By construction, the missing event is not a randomly missing event: it is, according to the semantic ontology, \textit{semantically related to another event in that sequence}. To compare our model with SSDVAE, we have trained SSDVAE with the same data and evaluated with the same masked input and full event regeneration.
%We have passed event tuples to the model but this time instead of passing all frames, we deleted one event and expect from the model to regenerate the whole event sequence from that. 

\begin{figure}[t]
\includegraphics{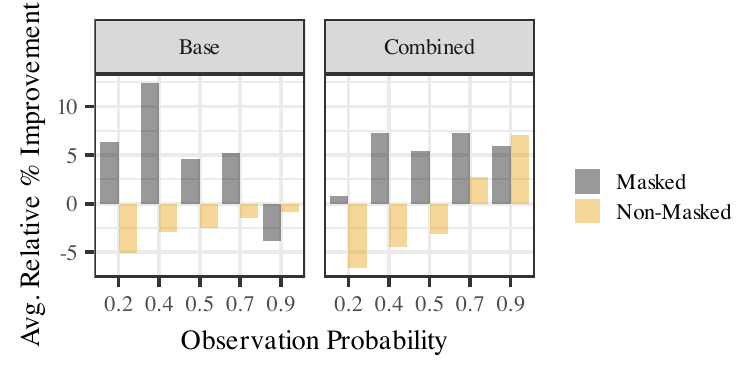}
\caption{Relative perplexity \textit{improvement} of the scenario-based model vs. the grouped model; higher is better. %We compare masked (grey) vs. non-masked (orange) event tuple generation. 
The scenario model improves across observation levels when important events are missing.}
\label{fig:grp-scn-factor-by-layer}
\end{figure}

Given their strong performance, we examine the grouped and scenario-based models. %
Results, averaged across three seeds, are in \cref{exp_masked_ppl}: \textit{grp} is the model with a group of FrameNet relations, \textit{scn} is the model with scenario sub-frames and \textit{SSDVAE} is the SSDVAE model with different evaluation. %From the results, we can see that having an extra layer of latent information is helpful for \textit{grp} model when the full event sequence is available. However, 
To show the consistent benefits of our approach, we report results computed just from the base decoder, just from the compression decoder, and from a score combined from both the base and compression decoders. %
When an important event is masked, the scenario-based model nearly always outperforms the grouped model across observation levels. Our model can leverage training time scenario-related frame associations to better predict a missing event. Also, for all observation probabilities, both of our model's (\textit{grp} \& \textit{scn}) individual and combined layer outperform SSDVAE. We suspect this is because  SSDVAE does not have a hierarchical abstraction mechanism, so when one event is not present, the related frame is also missing. This shows the capability of the hierarchical structure of our model to understand and encapsulate the abstract meaning of an event sequence. %
%the  \textit{scn} model is based on the scenario sub-frames which can identify connected events. %This experiment confirms the robustness of our scenario subframes to identify connected events and on providing more abstract generalization. 

It is not surprising that the base layer, with more feedback during training and greater representational capacity, is a better language model than the compression layer on its own. Still, the compression layer provides active benefits: we summarize the relative improvement of the scenario-based model over the grouped model in \cref{fig:grp-scn-factor-by-layer}. We compute this from just the base layer, or from both the base and compression layers. A positive number means that the scenario-based model was better able to (re)generate a full event sequence compared to the grouped model. Except for very high observation probability on the base layer, the \textbf{scenario-based model consistently outperformed the grouped one when semantically-relevant events were missing}. %This shows the capability of our scenario-based model to generate the full event sequences from masked input event when there are fewer annotations available. 
The grouped model, which covers multiple frame relations, can better model sequences when events are not missing. %\footnote{The grouped model had 4x more frame types available as guidance to the compression layer, vs. the scenario model.} 
While this may seem intuitive, notice how using the compression layer is able to reverse this pattern and let the scenario-based model outperform the grouped one, highlighting the benefit that the compression layer can bring.

% Contrastive learning
\subsection{Improved Event Similarity}
\label{sec:sentence-embeddings}
We have shown that both structural and semantic hierarchy can be beneficial when predicting the next event in a sequence, ``rolling out'' a longer sequence from an initial seed, and accounting for semantically missing events. %
In our final experiment, we use the latent frame representation to improve the overall event representation. We evaluate on three similarity datasets, comparing to the state-of-the-art~\citep{gao-etal-2022-improving}. 
In two of the datasets, there are two event pairs and the task is to determine which pair is more similar (measured by accuracy); the third involves scalar human assessment scores for how related two events are (Spearman correlation). %
Data are only for evaluation, and all training is done as ``pre-training.''
%In this experiment, we explored a different path beside the language modeling task, where we try to find out if our hierarchical model can be used to get efficient sentence embeddings which later can be used in downstream tasks. 
As such, our experiments demonstrate the ability to capture semantic information in our latent variable representation, and to perform in an evaluation-only (zero-shot) prediction of semantically-related events.

\citet{gao-etal-2022-improving} presents SWCC, a \textbf{s}imultaneous, \textbf{w}eakly supervised, \textbf{c}ontrastive learning and \textbf{c}lustering framework for event representation learning. They combine a clustering loss with the popular contrastive learning approach of InfoNCE \citep{oord2018representation}. Every ``query'' point $x$ (an event tuple) has positive (similar) instances $z_1, ..., z_R$, and negative (dissimilar) instances $z_{R+1}, ..., z_{S}$. Using a temperature-annealed similarity function on model-computed embeddings, e.g., cosine similarity on embeddings from a LLM, a probability distribution is computed over the positive and negatives (conditioned on the query). Average cross-entropy is optimized to predict the positive vs. negative instances. % 

This contrastive loss nicely augments our model's existing training objective from \cref{training}. %
We pre-train our hierarchical model on the same partially observable frame-annotated data from \cref{sec:experimental_setup}, using that model to extract a representation for an event, and computing the cosine similarity between two representations. %
We form a representation by concatenating the decoder's final token embedding and the latent frames from the compression layer. %
To prevent frame representations overfitting to the predicates, rather than arguments, we applied a predicate-specific dropout of 70\% on the encoder. %The main idea was to use the compressed frame and sentence embedding to cluster positive and negative events. 
%Dev experiments showed that predicate dropout rate of 70\% give the best result compared to others. 
Our hierarchical model provides a straightforward way to adopt contrastive loss; this hierarchical nature is not explicit in SSDVAE or HAQUE. Adapting these approaches to the contrastive learning setup is beyond the scope of our work.

%We used the neighboring events as positive data and in batch events as negative data points in our contrastive loss. We have used three similarity evaluation metrics same as \cite{gao-etal-2022-improving} to show our model's compression capability in sentence representation.

%One of the key differences between our model and SWCC is that they have used a weighted combination of contrastive loss, clustering loss and masked language model loss. In our model, we have used a weighted combination of reconstruction loss, frame classification loss, KL-divergence loss and contrastive loss. Also, we have used directed dropping out on predicates instead of whole embedding dropout in SWCC. Another key difference is that they have trained on 4.3M events, which is nearly double our dataset size (2.3M events).

% \begin{table}[t]
% \centering
% \resizebox{\columnwidth}{!}{%
% \begin{tabular}{c|c|c|c}
% \hline
% \multirow{2}{*}{Model} & \multicolumn{2}{c}{Hard Similarity (Accuracy \%)} & \multirow{2}{*}{\begin{tabular}[c]{@{}c@{}}Transitive Score\\ Similarity\end{tabular}} \\
%      & Original & Extended &       \\ \hline
% No InfoNCE, reconstruction & 57.39 & 58.20 & -0.02 \\
% InfoNCE, no reconstruction & 59.13    & 73.5   & 0.63 \\
% InfoNCE, reconstruction & 66.09	& 67.6 & 0.75 \\ \hline
% \end{tabular}%
% }
% \caption{Ablation study on Similarity Tasks — Experiment \cref{sec:sentence-embeddings}}
% \label{exp_swcc_abl}
% \end{table

\begin{table}[t]
\centering
\resizebox{\columnwidth}{!}{%
\begin{tabular}{c|c|c|c}
\hline
\multirow{2}{*}{Model} & \multicolumn{2}{c}{Hard Similarity (Accuracy \%)} & \multirow{2}{*}{\begin{tabular}[c]{@{}c@{}}Transitive Score\\ Similarity\end{tabular}} \\
     & Original & Extended &       \\ \hline
SWCC (16) & 78.91 $\pm$ 1.31    & 69.2 $\pm$ 0.93    & 0.82 $\pm$ 0 \\
SWCC (256) & 81.09 $\pm$ 0.43    & 72.55 $\pm$ 1.53    & \textbf{0.82 $\pm$ 0} \\
Ours & \textbf{83.26 $\pm$ 2.29}    & \textbf{78.63 $\pm$ 2.95}    & 0.77 $\pm$ 0.04 \\ \hline
\end{tabular}%
}
\caption{Evaluation on Similarity Tasks. SWCC (256) are \citeauthor{gao-etal-2022-improving}'s reported results, using a batch size of 256. Given the importance that batch size can have with contrastive learning, we ran \citeauthor{gao-etal-2022-improving}'s model with a batch size 16 (the same batch size of our model). We report this as SWCC (16). See \cref{sec:sentence-embeddings}.}
\label{exp_swcc}
\end{table}

% https://docs.google.com/document/d/1gr77mGwj_F1ILn6Y-1sc1wJ3hykTI12WWFkRyelosws/edit#
% \begin{table}[t]
% \centering
% \resizebox{\columnwidth}{!}{%
% \begin{tabular}{c|c|c|c}
% \hline
% \multirow{2}{*}{Model} & \multicolumn{2}{c}{Hard Similarity (Accuracy \%)} & \multirow{2}{*}{\begin{tabular}[c]{@{}c@{}}Transitive Score\\ Similarity\end{tabular}} \\
%      & Original & Extended &       \\ \hline
% SWCC (16) (contrastive + MLM) & 78.91    & 69.2    & 0.82 \\
% SWCC (16) (contrastive) & 74.78 \textcolor{red}{-4.13} & 67.60 \textcolor{red}{-1.6} & 0.81 \textcolor{red}{-0.01} \\
% SWCC (16) (MLM) & 6.09 \textcolor{red}{-72.82}    & 7.50 \textcolor{red}{-61.7}  & 0.38 \textcolor{red}{-0.44} \\ \hline
% \end{tabular}%
% }
% \caption{\ForFFInline{This table is new}
% \ForDiptaInline{See email: do we have similar ablation studies for our approach? Work in Progress}
% Ablation study on Similarity Tasks — Experiment \cref{sec:sentence-embeddings}}
% \label{exp_swcc_abl}
% \end{table}

% https://docs.google.com/spreadsheets/d/1a95khvHujARJrvhCZ3a1IGhYsNjK20gTRs4jfDmJVMU/edit#gid=0
\begin{table}[t]
\centering
\resizebox{\columnwidth}{!}{%
\begin{tabular}{c|c|c|c|c}
\hline
& \multirow{2}{*}{Training Variant} & \multicolumn{2}{c}{Hard Similarity (Accuracy \%)} & \multirow{2}{*}{\begin{tabular}[c]{@{}c@{}}Transitive Score\\ Similarity\end{tabular}} \\
  &   & Original & Extended &       \\ \hline
\multirow{3}{*}{Ours (16)}   & Contrastive + LM      & 83.26 $\pm$ 2.29      & 78.63 $\pm$ 2.95      & 0.77 $\pm$ 0.04   \\
                        & Contrastive only      & 67.18 $\pm$ 1.79      & 72.75 $\pm$ 2.06      & 0.72 $\pm$ 0.02   \\
                        & LM only               & 67.83 $\pm$ 14.39     & 62.15 $\pm$ 16.52     & 0.56 $\pm$ 0.04   \\ \hline
\multirow{3}{*}{SWCC (16)}   & Contrastive + MLM     & 78.91 $\pm$ 1.31      & 69.2 $\pm$ 0.93       & 0.82 $\pm$ 0      \\
                        & Contrastive only      & 78.48 $\pm$ 0.83      & 67.33 $\pm$ 0.19      & 0.78 $\pm$ 0.05   \\
                        & MLM only              & 25.87 $\pm$ 1.31      & 16.78 $\pm$ 0.7       & 0.55 $\pm$ 0.04   \\ \hline
\end{tabular}%
% \textcolor{red}{-16.08} \textcolor{red}{-6.25} \textcolor{red}{-0.05} 
% \textcolor{red}{-15.43} \textcolor{red}{-16.48} \textcolor{red}{-0.21}
}
\caption{Ablation study of our model and SWCC. %
}
\label{exp_swcc_abl}
\end{table}

Our results are in \cref{exp_swcc}. We have run SWCC with a batch size of 16, which is the same as ours. Our model surpasses SWCC on two of the tasks, showing it is not only capable of event language modeling but also capable of generating better event representations. We have also run an ablation study on SWCC and our model; the results are on \cref{exp_swcc_abl}. The results show that neither contrastive nor LM/MLM loss are as strong as both together. We see that the LM component in our approach is important to overall performance.

\section{Conclusion}
We have presented a hierarchical event model that accounts for both structural and ontological hierarchy across an event sequence. We use automatically extracted semantic frames to guide the first level of concept, and then use FrameNet relations to guide abstraction and generalization. We showed improvements across multiple tasks and evaluation measures within event modeling. We showed improvements in next event prediction, longer range event prediction, missing event regeneration, and event similarity. We believe that future work can use this abstraction concept for summarization, topic modeling, or other downstream tasks. % i.e., clustering of related events.

\section{Limitations}
Our approach enables modeling observed event sequences through the lens of a structured semantic ontology. %
Though our models have shown superior performance to leverage event frames, they still suffer from the bottleneck of the information passed to the compression layer. %
Additionally,  while these resources do exist, their coverage is not universal, and have historically been developed for English. Our experiments reflect this.

While the observance of frames is not, strictly speaking, a requirement of our model, our experiments focused on those cases when such an ontology is available during training.

Throughout our experiments, we use pretrained models/embeddings. We do not attempt to control or mitigate any biases these may exhibit or propagate.

Our work does not involve human subjects research, data annotation, or representation/analysis of potentially sensitive characteristics. As such, while we believe the direct \textit{potential risks} of our approach are minimal we acknowledge that the joint use of pretrained models and structured semantic ontologies could result in undesired or biased semantic associations.

\section*{Acknowledgments}
We would like to thank the anonymous reviewers for their comments, questions, and suggestions. %
This material is based in part upon work supported by the National Science Foundation under Grant No. IIS-2024878. %
Some experiments were conducted on the UMBC HPCF, supported by the National Science Foundation under Grant No. CNS-1920079. %
This material is also based on research that is in part supported by the Army Research Laboratory, Grant No. W911NF2120076, and by the Air Force Research Laboratory (AFRL), DARPA, for the KAIROS program under agreement number FA8750-19-2-1003. The U.S. Government is authorized to reproduce and distribute reprints for Governmental purposes notwithstanding any copyright notation thereon. The views and conclusions contained herein are those of the authors and should not be interpreted as necessarily representing the official policies or endorsements, either express or implied, of the Air Force Research Laboratory (AFRL), DARPA, or the U.S. Government. %

\bibliographystyle{acl_natbib}
\bibliography{anthology,custom}

% \clearpage
\appendix

\section{Additional Model and Implementation Details}

\subsection{Model Details} \label{sec:more-model-details}
For our input data, events are separated by a \textit{<TUP>} token, and in case of missing values in an event frame, is replaced with a special \textit{<NOFRAME>} token. 

As mentioned in the main paper, like any auto-regressive model, previously generated decoder output and previous input texts are given as input to the decoder. An attention module is used to find the important words from the given latent embeddings predicted by encoder. %
Each layer tries to reconstruct the input text, and loss was generated individually for each layer, which then accumulated and back-propagated through the whole model, updating the model parameters.

\subsection{Implementation Details} \label{sec:implementation-details}

The values of $\gamma_1 $ and $\gamma_2 $ are set to 0.1 by experimenting on the validation set. 2 Gumbel-softmax samples are used to average the encoder. We use the Adam~\citep{kingma2014adam} optimizer with a learning rate of 0.001. A batch size of 64 has been used with a gradient accumulation of 8. Early stopping has been used with patience of 10 on the validation perplexity score. 

For comparability, our core event modeling results use recurrent encoders and decoders. We use pretrained Glove-300 embeddings to represent each lexical item in an event tuple. An embedding size of 500 has been used for frame embeddings. Two layers of bidirectional GRU have been used for the encoder, and two layers of uni-directional GRU have been used for the decoder. Both are used with 512 hidden sizes. Gradient clipping of 5.0 has been used to prevent gradient exploding. 0.5 has been used as the Gumbel-softmax temperature. 

Similarly, our experiments involving event similarity (\cref{sec:sentence-embeddings}) use BART~\cite{lewis2019bart} as our encoder and decoder module. 

Across our experiments, we have used NVIDIA RTX 2080Ti or NVIDIA RTX 6000 for training. It takes around 16 hours to train with our current batch size on our dataset.

\subsection{Loss Formulation}\label{sec:loss-formulation}

In constructing our training loss function, we take inspiration from the methodology outlined in the study conducted by \citep{rezaee-ferraro-2021-event}.
 However, our model differs in that it incorporates two hidden layers, as opposed to the single latent layer utilized in the aforementioned study.
 Each layer
 we calculate the loss for both layers individually. This is done by allowing each layer $j$, to reconstruct the input text using its own latent variables, $L_{r_j}$. To prevent overfitting, we incorporate $\text{KL}$ terms in our loss function denoted as $\mathcal{L}_{\text{KL}_j}$. Additionally, for the base layer we  include a classification term, designated as $\mathcal{L}_c$.
% The base layer receives input that includes partially observed frames, determined by a specified observation probability. The first layer encoder utilizes these partially observed frames as guidance to encode the input sequence, resulting in a latent variable representation, denoted as $(f_{i})$. This predicted latent variable $(f_i)$ is then passed through the decoder to regenerate the text. The second layer encoder also receives input, in the form of the predicted frames from the first layer and their parent frames, and encodes them into a smaller number of latent variables, $(h_j)$, which are subsequently used in the decoder. 

\begin{equation}
    \begin{aligned}
        \mathcal{L} &= 
        \underbrace{\alpha_1 * \mathcal{L}_{r_1} + \alpha_2 * \mathcal{L}_{r_2}}_ 
        {\text{Text Reconstruction}}
        \\
        &+ \: 
        \underbrace{
        \beta_1 * \mathcal{L}_{\text{KL}_1} + \beta_2 
        * \mathcal{L}_{\text{KL}_2}
        }_
        {\text{Regularization}}
        \\
        &+ \: 
        \underbrace{
        \gamma * \mathcal{L}_{c}. 
        }_
        {\text{Observed Frame Classification}}
                    %%+ \gamma_2 * \mathcal{L}_{\gamma_2}
    \end{aligned}
\end{equation}

The reconstruction and KL losses depend on the random variables inferred at each level: for the base level ($j=1$), the losses depends on the frames sampled at the base level $f_1, \ldots, f_n$, while the compression losses ($j=2$) depend on $h_1, \ldots, h_M$). %
Our latent variable model learns a variational distribution $q$, from which it can infer appropriate values for $f_i$ and $h_j$. With this, we compute
\begin{align}
\mathcal{L}_{r_1} & = \mathbb{E}_{q(f_1,\ldots,f_N)}[\log p(x | f_1, \ldots, f_N)] \\
\mathcal{L}_{r_2} & = \mathbb{E}_{q(h_1,\ldots,h_M)}[\log p(x | h_1, \ldots, h_M)] \\
\mathcal{L}_{\text{KL}_1} & = \mathbb{E}_{q(f_1,\ldots,f_N)}[\log p(f_1, \ldots, f_N)] \\
\mathcal{L}_{\text{KL}_2} & = \mathbb{E}_{q(h_1,\ldots,h_M)}[\log p(h_1, \ldots, h_M)] \\
\mathcal{L}_{c} & = -\sum_{i=1: f^*_i\text{ is obs.}}^N \log q(f^*_i | f_{i-1}).
\end{align}
In $\mathcal{L}_c$, note that $f^*_i$ represents the correct value of the $i$th frame. The reconstruction and frame classification losses can be computed via a cross-entropy loss (per output token for the reconstruction losses, and per predicted frame in the frame classification loss).

\section{Additional Results} 

% exp 3.5
\subsection{Is Frame Inheritance Sufficient?} \label{exp_3.5-app}
The detailed results for the experiment described in \cref{exp_3.5} are reported in \cref{exp_3.5_12_ppl} (Perplexity Score) and \cref{exp_3.5_12_winv} (INC).

Detailed per-layer perplexity is reported in \cref{exp_3.5_12_ppl}, augmenting the results in \cref{exp_3.5_12_combined}. Our model's base layer perplexity consistently outperformed the other models. However, the perplexity of the compression layer was higher. %
This suggests that while incorporating hierarchical layers or knowledge may not be sufficient for generating the event sequence, it provides useful, less-than-full supervised feedback to the base layer.

For INC, we look to the lexical variant, where our model's base layer outperforms the previous result with having the best of all the observation probabilities. However, the results for the compression layer underperformed the inferred variant, indicating that incorporating lexical signals may have a negative impact on the performance of the generation model. Overall, this suggests that the \textbf{inferred frames and ontological relations from the base layer are important for hierarchical modeling}.

\begin{figure}[t]
\includegraphics{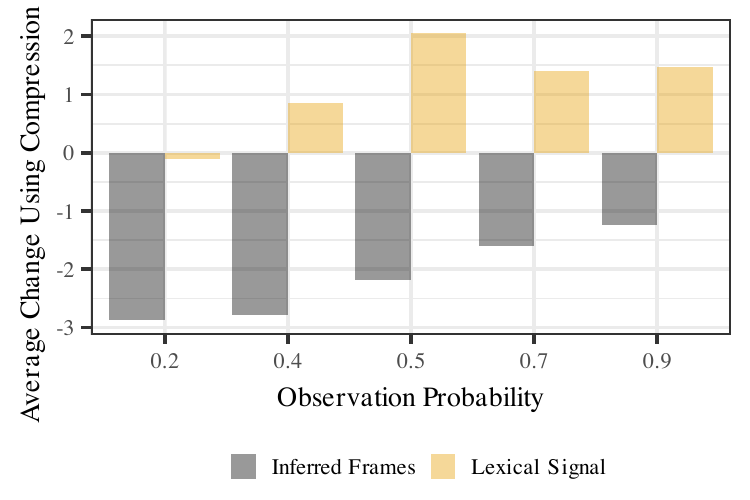}
\caption{Average Change in INC score from combined layer to base layer, where a negative score means the base layer was better than the combined one and vice versa. The gray and orange bars indicate whether the input to the compression layer is inferred frames or lexical signal, respectively. In all cases, inferred frames has a better effect on base layer and lexical signal has improved combined layer's performance.}
\label{fig:improvement-from-compression}
\end{figure}

We have reported an average change in the INC score of the base layer over the combined layer on \cref{fig:improvement-from-compression}. The gray and orange bars represent the two variants: inferred frames and lexical signal, respectively. Each bar is the average of the score change from the combined layer to the base layer (combined layer score – base layer score). Here, a negative score means that the base layer is better than the combined one. This figure shows if the use of compression layer has a positive impact on the INC score or not. First, for the inferred frames, the addition of a compression layer has improved the INC score by an effective margin on the base layer. This shows that the semantic frames have helped the model's base layer to understand the process better. On the other hand, for the lexical signal, the combined layer has a better INC score. This shows that having the lexical signal on the compression layer has a better and equal effect on both layers. In conclusion, the addition of a compression layer improves the model's capability of understanding event sequences and generalizing.

\begin{table}[t]
\resizebox{\columnwidth}{!}{%
\begin{tabular}{c|c|ccc}
\hline
\multirow{2}{*}{Model}  & \multirow{2}{*}{$\epsilon$}   & \multicolumn{3}{c}{Perplexity (Test Data)} \\ 
                        &                      & Base     & Compression    & Total   \\ \hline
HAQAE                   & -                    & -        & -              & 21.38 $\pm$ 0.25   \\ \hdashline
SSDVAE                  & \multirow{3}{*}{0.9} & -        & -              & 19.84 $\pm$ 0.52   \\ 
ours: inf. frame                 &                      & \textit{19.39 $\pm$ 0.3}    & 26.52 $\pm$ 0.55          & 22.68 $\pm$ 0.41   \\
ours: lexical                   &                      & \textit{\textbf{19.12 $\pm$ 0.53}}    & 31.43 $\pm$ 1.1          & 24.51 $\pm$ 0.39   \\ \hdashline

SSDVAE                  & \multirow{3}{*}{0.7} & -        & -              & 21.19 $\pm$ 0.76   \\ 
ours: inf. frame                 &                      & \textit{20.26 $\pm$ 1.36}    & 27.45 $\pm$ 0.5          & 23.57 $\pm$ 0.84   \\ 
ours: lexical                   &                      & \textit{21.52 $\pm$ 1.48}    & 35.19 $\pm$ 0.95          & 27.5 $\pm$ 0.93   \\ \hdashline

SSDVAE                  & \multirow{3}{*}{0.5} & -        & -              & 31.11 $\pm$ 0.85   \\ 
ours: inf. frame                 &                      & \textit{22.16 $\pm$ 1.62}    & 32.59 $\pm$ 2.86          & 26.62 $\pm$ 2.13   \\
ours: lexical                   &                      & \textit{25.02 $\pm$ 1.31}    & 39.44 $\pm$ 0.44          & 31.41 $\pm$ 0.77   \\ \hdashline

SSDVAE                  & \multirow{3}{*}{0.4} & -        & -              & 33.12 $\pm$ 0.54   \\ 
ours: inf. frame                 &                      & \textit{24.02 $\pm$ 1.28}    & 32.82 $\pm$ 1.44          & 28.07 $\pm$ 1.24   \\
ours: lexical                   &                      & \textit{27.06 $\pm$ 0.94}    & 40.46 $\pm$ 2.74          & 33.05 $\pm$ 0.56   \\ \hdashline

SSDVAE                  & \multirow{3}{*}{0.2} & -        & -              & 33.31 $\pm$ 0.63   \\ 
ours: inf. frame                 &                      & \textit{30.15 $\pm$ 2.73}    & 34.81 $\pm$ 2.81          & 32.84 $\pm$ 1.84   \\
ours: lexical                   &                      & \textit{33.6 $\pm$ 1.84}    & 44.64 $\pm$ 1.44          & 38.72 $\pm$ 1.59   \\ \hline
\end{tabular}%
}
\caption{Per-word perplexity for test data (lower is better). For each observation probability ($\epsilon$), the best perplexity is in \textit{italic} form. The best of all of them is \textbf{bold} form. See \cref{exp_3.5-app}} %, \cref{exp_3.12}}
\label{exp_3.5_12_ppl}
\end{table}

\begin{table}[t]
\resizebox{\columnwidth}{!}{%
\begin{tabular}{c|c|ccc}
\hline
\multirow{2}{*}{Model} & \multirow{2}{*}{$\epsilon$}   & \multicolumn{3}{c}{Wikipedia INC (Test Data)} \\ 
                       &                      & Base     & Compression    & Total   \\ \hline
HAQAE                  & -                    & -        & -              & 24.88 $\pm$ 1.35   \\ \hdashline

SSDVAE                 & \multirow{3}{*}{0.9} & -        & -              & 35.56 $\pm$ 1.70   \\ 
ours: inf. frame                   &                      & 41.35 $\pm$ 4.25    & 27.25 $\pm$ 1.02          & 40.11 $\pm$ 1.88   \\
ours: lexical                   &                      & 41.35 $\pm$ 3.19    & 35.41 $\pm$ 2.56          & \textit{42.83 $\pm$ 1.47}   \\ \hdashline

SSDVAE                 & \multirow{3}{*}{0.7} & -        & -              & \textit{39.08 $\pm$ 1.55}   \\ 
ours: inf. frame                   &                      & 35.86 $\pm$ 3.43    & 26.31 $\pm$ 2.92          & 34.26 $\pm$ 3.43   \\
ours: lexical                   &                      & 35.61 $\pm$ 4.72    & 32.68 $\pm$ 6.12          & 37.01 $\pm$ 6.59   \\ \hdashline

SSDVAE                 & \multirow{3}{*}{0.5} & -        & -              & \textit{40.18 $\pm$ 0.90}   \\ 
ours: inf. frame                   &                      & 37.3 $\pm$ 3.33    & 23.61 $\pm$ 1.34          & 35.13 $\pm$ 3.01   \\
ours: lexical                   &                      & 37.8 $\pm$ 3    & 37.11 $\pm$ 3.14          & 39.85 $\pm$ 3.01   \\ \hdashline

SSDVAE                 & \multirow{3}{*}{0.4} & -        & -              & \textit{47.88 $\pm$ 3.59}   \\ 
ours: inf. frame                   &                      & 43.25 $\pm$ 4.97    & 23.65 $\pm$ 1.34          & 40.46 $\pm$ 4.71   \\
ours: lexical                   &                      & 39.2 $\pm$ 1.23    & 34.79 $\pm$ 4.75          & 40.06 $\pm$ 2   \\ \hdashline

SSDVAE                 & \multirow{3}{*}{0.2} & -        & -              & 44.38 $\pm$ 2.10   \\ 
ours: inf. frame                   &                      & \textit{\textbf{49.53 $\pm$ 1.56}}    & 25.15 $\pm$ 4.34          & 46.65 $\pm$ 1.55   \\
ours: lexical                   &                      & 46.53 $\pm$ 2.84    & 37.55 $\pm$ 2.8          & 46.41 $\pm$ 3.71   \\ \hline
\end{tabular}%
}
\caption{Wikipedia Inverse Narrative Cloze Score for test data (higher is better). For each observation probability ($\epsilon$), the best score is in \textit{italic} form. The best of all of them is \textbf{bold} form. See \cref{exp_3.5-app}} %, \cref{exp_3.12}}
\label{exp_3.5_12_winv}
\end{table}

\subsection{The Effect of Individual Frame Relations}  \label{sec:frame-relations-appendix}
The detailed results for the experiment described in \cref{exp_3.10} are reported on \cref{exp_3.10_ppl} (Perplexity Score) and \cref{exp_3.10_winv} (Wikipedia Inverse Narrative Score).

\begin{table}[t]
\resizebox{\columnwidth}{!}{%
\begin{tabular}{c|c|c|ccc}
\hline
\multirow{2}{*}{Model} & \multirow{2}{*}{Frame Relation} &  \multirow{2}{*}{$\epsilon$}   & \multicolumn{3}{c}{Perplexity (Test Data)} \\ 
                       &                        &                  & Base     & Compression    & Total   \\ \hline
HAQAE                  & - & -                    & -        & -                    & 21.38 $\pm$ 0.25   \\ \hdashline

SSDVAE                  & - & \multirow{10}{*}{0.9}     & -        & -              & 19.84 $\pm$ 0.52   \\ 
\multirow{9}{*}{ours}   & Using &                       & 19.39 $\pm$ 0.51    & 25.34 $\pm$ 0.22          & 22.16 $\pm$ 0.37   \\ 
                        & Precedes &                    & 19.57 $\pm$ 0.58    & 25.83 $\pm$ 0.25          & 22.48 $\pm$ 0.25   \\
                        & Metaphor &                    & 19.62 $\pm$ 0.75    & 25.21 $\pm$ 0.49          & 22.24 $\pm$ 0.63   \\
                        & See\_also &                   & 19.55 $\pm$ 0.72    & 25.71 $\pm$ 0.39          & 22.42 $\pm$ 0.54   \\
                        & Causative\_of &               & 19.42 $\pm$ 0.57    & 25.75 $\pm$ 0.46          & 22.36 $\pm$ 0.53   \\
                        & Inchoative\_of &              & 19.28 $\pm$ 0.32    & 26.01 $\pm$ 0.85          & 22.39 $\pm$ 0.52   \\
                        & Perspective\_on &             & 19.76 $\pm$ 0.97    & 25.64 $\pm$ 0.57          & 22.5 $\pm$ 0.75   \\
                        & Subframe &                    & \textbf{\textit{18.91 $\pm$ 0.15}}    & 26.03 $\pm$ 0.42          & 22.19 $\pm$ 0.27   \\
                        & ReFraming\_Mapping &          & 19.56 $\pm$ 0.94    & 26.63 $\pm$ 1.81          & 22.81 $\pm$ 0.62   \\ \hdashline

% SSDVAE                 & \multirow{3}{*}{0.7} & -        & -              & 21.19   \\ 
% ours+frame              &                      & 18.22    & 30.00          & 23.38   \\ 
% ours+evt                &                      & 19.98    & 30.30          & 24.60   \\ \hdashline

% SSDVAE                 & \multirow{3}{*}{0.5} & -        & -              & 31.11   \\ 
% ours+frame              &                      & 20.72    & 32.26          & 25.85   \\ 
% ours+evt                &                      & 22.83    & 36.17          & 28.94   \\ \hdashline

% SSDVAE                 & \multirow{3}{*}{0.4} & -        & -              & 33.12   \\ 
% ours+frame              &                      & 23.18    & 23.97          & 27.27   \\ 
% ours+evt                &                      & 28.13    & 37.32          & 32.40   \\ \hdashline

SSDVAE                 & - & \multirow{10}{*}{0.2} & -        & -                   & 33.31 $\pm$ 0.63    \\ 
\multirow{9}{*}{ours}   & Using &                       & 31.37 $\pm$ 2.08    & 38.55 $\pm$ 5.72          & 34.72 $\pm$ 3.23   \\ 
                        & Precedes &                    & 32.62 $\pm$ 1.65    & 45.33 $\pm$ 0.74          & 38.45 $\pm$ 1.25   \\
                        & Metaphor &                    & 32.92 $\pm$ 2.08    & 42.07 $\pm$ 5.83          & 37.18 $\pm$ 3.5   \\
                        & See\_also &                   & 31.83 $\pm$ 2.78    & 41.78 $\pm$ 5.55          & 36.44 $\pm$ 3.79   \\
                        & Causative\_of &               & 31.82 $\pm$ 3    & 40.01 $\pm$ 6.23          & 35.67 $\pm$ 4.41   \\
                        & Inchoative\_of &              & 32.65 $\pm$ 1.4    & 42.42 $\pm$ 3.55          & 37.21 $\pm$ 2.21   \\
                        & Perspective\_on &             & 33.2 $\pm$ 1.47    & 44.18 $\pm$ 1.26          & 38.28 $\pm$ 0.34   \\
                        & Subframe &                    & 32.78 $\pm$ 2.09    & 45.25 $\pm$ 0.7          & 38.51 $\pm$ 1.52   \\
                        & ReFraming\_Mapping &          & \textit{31.34 $\pm$ 2.76}    & 36.57 $\pm$ 2.9          & 34.06 $\pm$ 3.15   \\ \hline
\end{tabular}%
}
\caption{Per-word perplexity for test data (lower is better). For each observation probability ($\epsilon$), the best perplexity is in \textit{italic} form. The best of all of them is \textbf{bold} form. See \cref{sec:frame-relations-appendix}}
\label{exp_3.10_ppl}
\end{table}

\begin{table}[t]
\resizebox{\columnwidth}{!}{%
\begin{tabular}{c|c|c|ccc}
\hline
\multirow{2}{*}{Model} & \multirow{2}{*}{Frame Relation} & \multirow{2}{*}{$\epsilon$}   & \multicolumn{3}{c}{Wikipedia INC (Test Data)} \\ 
                       & - &                     & Base        & Compression           & Total   \\ \hline
HAQAE                  & -                    & -           & -            & -         & 24.88 $\pm$ 1.35   \\ \hdashline

SSDVAE                 & - & \multirow{10}{*}{0.9} & -        & -                   & 35.56 $\pm$ 1.70   \\ 
\multirow{9}{*}{ours}   & Using &                       & 43.23 $\pm$ 2.51    & 26.68 $\pm$ 0.63          & 40.92 $\pm$ 1.85   \\ 
                        & Precedes &                    & 41.43 $\pm$ 3.02    & 26.38 $\pm$ 1.51          & 40.03 $\pm$ 1.66   \\
                        & Metaphor &                    & 41.92 $\pm$ 3.93    & 24.22 $\pm$ 1.53          & 38.8 $\pm$ 2.17   \\
                        & See\_also &                   & 42.67 $\pm$ 1.49    & 27.08 $\pm$ 0.24          & 41.13 $\pm$ 0.81   \\
                        & Causative\_of &               & 41.38 $\pm$ 2.23    & 26.3 $\pm$ 1.05          & 40.47 $\pm$ 1.79   \\
                        & Inchoative\_of &              & 41.35 $\pm$ 3.47    & 26.67 $\pm$ 1.33          & 40 $\pm$ 2.34   \\
                        & Perspective\_on &             & 40.53 $\pm$ 2.04    & 26.38 $\pm$ 0.67          & 39.55 $\pm$ 1.75   \\
                        & Subframe &                    & 40.35 $\pm$ 2.91    & 25.7 $\pm$ 0.48          & 38.42 $\pm$ 2.32   \\
                        & ReFraming\_Mapping &          & \textit{43.8 $\pm$ 4.02}    & 26.7 $\pm$ 1.21          & 42.15 $\pm$ 3.19   \\ \hdashline

% SSDVAE                 & \multirow{2}{*}{0.7} & -        & -              & \textit{39.08}   \\ 
% ours+frame              &                      & 32.05    & 26.10          & 32.40   \\ 
% ours+evt                &                      & 33.20    & 38.50          & 37.80   \\ \hdashline

% SSDVAE                 & \multirow{2}{*}{0.5} & -        & -              & \textit{40.18}   \\ 
% ours+frame              &                      & \textit{40.75}    & 24.75          & 38.55   \\ 
% ours+evt                &                      & 34.90    & 40.65          & 40.55   \\ \hdashline

% SSDVAE                 & \multirow{2}{*}{0.4} & -        & -              & \textit{47.88}   \\ 
% ours+frame              &                      & 44.35    & 25.00          & 42.95   \\ 
% ours+evt                &                      & 41.40    & 41.75          & 45.00   \\ \hdashline

SSDVAE                 & - & \multirow{10}{*}{0.2} & -        & -                   & 44.38 $\pm$ 2.10   \\ 
\multirow{9}{*}{ours}   & Using &                       & \textbf{\textit{49.72 $\pm$ 1.73}}    & 21.77 $\pm$ 1.1          & 45.93 $\pm$ 1.62   \\ 
                        & Precedes &                    & 47.92 $\pm$ 2.25    & 20.67 $\pm$ 0.29          & 42.72 $\pm$ 1.58   \\
                        & Metaphor &                    & 47.25 $\pm$ 3.81    & 21.12 $\pm$ 0.95          & 42.77 $\pm$ 3.27   \\
                        & See\_also &                   & 47.77 $\pm$ 3.61    & 21.2 $\pm$ 1.15          & 43.72 $\pm$ 2.78   \\
                        & Causative\_of &               & 49.85 $\pm$ 0.84    & 21.5 $\pm$ 2.41          & 45.45 $\pm$ 2.03   \\
                        & Inchoative\_of &              & 48.03 $\pm$ 3.35    & 21 $\pm$ 0.74          & 43.95 $\pm$ 2.61   \\
                        & Perspective\_on &             & 47.85 $\pm$ 3.53    & 20.42 $\pm$ 0.3          & 43.08 $\pm$ 3.12   \\
                        & Subframe &                    & 47.88 $\pm$ 3.31    & 20.33 $\pm$ 0.52          & 42.38 $\pm$ 1.86   \\
                        & ReFraming\_Mapping &          & 49.05 $\pm$ 1.54    & 22.23 $\pm$ 0.58          & 45.45 $\pm$ 0.44   \\ \hline
\end{tabular}%
}
\caption{Wikipedia Inverse Narrative Cloze Score for test data (higher is better). For each observation probability ($\epsilon$), the best score is in \textit{italic} form. The best of all of them is \textbf{bold} form. See \cref{sec:frame-relations-appendix}}
\label{exp_3.10_winv}
\end{table}

\subsection{Are scenario subframes better than other frame properties?} \label{sec:exp_3.19_appendix}
The detailed results for the experiment reported in \cref{exp_3.10_combined} are shown in \cref{exp_3.19_ppl} (Perplexity Score) and \cref{exp_3.19_winv} (INC).

\begin{table}[t]
\resizebox{\columnwidth}{!}{%
\begin{tabular}{c|c|ccc}
\hline
\multirow{2}{*}{Model}  & \multirow{2}{*}{$\epsilon$}   & \multicolumn{3}{c}{Perplexity (Test Data)} \\ 
                        &                       & Base      & Compression       & Total   \\ \hline
HAQAE                   & -                     & -         & -                 & 21.38 $\pm$ 0.25   \\ \hdashline

SSDVAE                  & \multirow{2}{*}{0.9}  & -         & -                 & 19.84 $\pm$ 0.52   \\ 
\textit{scenario-only}                    &                       & \textit{18.81 $\pm$ 0.36}     & 25.61 $\pm$ 1.23             & 21.94 $\pm$ 0.5   \\ \hdashline

SSDVAE                  & \multirow{2}{*}{0.7}  & -         & -                 & 21.19 $\pm$ 0.76   \\ 
\textit{scenario-only}                    &                       & \textbf{\textit{18.75 $\pm$ 0.3}}     & 26.82 $\pm$ 0.47             & 22.42 $\pm$ 0.21   \\ \hdashline

SSDVAE                  & \multirow{2}{*}{0.5}  & -         & -                 & 31.11 $\pm$ 0.85   \\ 
\textit{scenario-only}                    &                       & \textit{23.79 $\pm$ 1.29}     & 31.43 $\pm$ 7.44             & 28.7 $\pm$ 2.04   \\ \hdashline

SSDVAE                  & \multirow{2}{*}{0.4}  & -         & -                 & 33.12 $\pm$ 0.54   \\ 
\textit{scenario-only}                    &                       & \textit{25.54 $\pm$ 2.34}     & 36.87 $\pm$ 6.01             & 30.63 $\pm$ 3.52   \\ \hdashline

SSDVAE                  & \multirow{2}{*}{0.2}  & -         & -                 & 33.31 $\pm$ 0.63   \\ 
\textit{scenario-only}                    &                       & \textit{32.01 $\pm$ 0.7}     & 45.28 $\pm$ 0.7             & 38.07 $\pm$ 0.55   \\ \hline
\end{tabular}%
}
\caption{Per-word perplexity for test data (lower is better). For each observation probability ($\epsilon$), the best perplexity is in \textit{italic} form. The best of all of them is \textbf{bold} form. See \cref{sec:exp_3.19_appendix}}
\label{exp_3.19_ppl}
\end{table}

\begin{table}[]
\resizebox{\columnwidth}{!}{%
\begin{tabular}{c|c|ccc}
\hline
\multirow{2}{*}{Model}  & \multirow{2}{*}{$\epsilon$}   & \multicolumn{3}{c}{Wikipedia INC (Test Data)} \\ 
                        &                       & Base      & Compression    & Total   \\ \hline
HAQAE                   & -                     & -         & -              & 24.88 $\pm$ 1.35   \\ \hdashline

SSDVAE                  & \multirow{2}{*}{0.9}  & -         & -              & 35.56 $\pm$ 1.70   \\ 
\textit{scenario-only}                    &                       & \textit{42.29 $\pm$ 1.79}     & 25.38 $\pm$ 1.84          & 39.86 $\pm$ 1.82   \\ \hdashline

SSDVAE                  & \multirow{2}{*}{0.7}  & -         & -              & \textit{39.08 $\pm$ 1.55}   \\ 
\textit{scenario-only}                    &                       & 38.79 $\pm$ 4.11     & 26.83 $\pm$ 7.32          & 32.91 $\pm$ 7.29   \\ \hdashline

SSDVAE                  & \multirow{2}{*}{0.5}  & -         & -              & \textit{40.18 $\pm$ 0.90}   \\ 
\textit{scenario-only}                    &                       & 37.59 $\pm$ 5.61     & 22.06 $\pm$ 1.01          & 35.59 $\pm$ 4.71   \\ \hdashline

SSDVAE                  & \multirow{2}{*}{0.4}  & -         & -              & \textit{47.88 $\pm$ 3.59}   \\ 
\textit{scenario-only}                    &                       & 40.91 $\pm$ 2.19     & 22.15 $\pm$ 1.37          & 37.99 $\pm$ 1.86   \\ \hdashline

SSDVAE                  & \multirow{2}{*}{0.2}  & -         & -              & 44.38 $\pm$ 2.10   \\ 
\textit{scenario-only}                    &                       & \textbf{\textit{48.1 $\pm$ 2.22}}     & 20.54 $\pm$ 0.1          & 43.3 $\pm$ 2.33   \\ \hline
\end{tabular}%
}
\caption{Wikipedia Inverse Narrative Cloze Score for test data (higher is better). For each observation probability ($\epsilon$), the best score is in \textit{italic} form. The best of all of them is \textbf{bold} form. See \cref{sec:exp_3.19_appendix}}
\label{exp_3.19_winv}
\end{table}

% exp 3.13
\subsection{The Effect of Grouping Frame Properties} \label{sec:exp_3.13-app}
The previous section showed that performance of our model can be further improved by using targeted frame relations. Here, we investigate whether grouping of different frame relations could have a more significant impact on generalization. 

Using \cref{exp_3.10}, we identified six frame-relations as the most important ones: Inheritance, Using, Precedes, Causative\_of, Inchoative\_of, and Subframe. We used this group of frame relations to extract the parent frames from the predicted frames of the base layer. With their parent frames, these frames were passed to the compression layer to learn to associate the semantically similar frames.

\begin{table}[t]
\resizebox{\columnwidth}{!}{%
\begin{tabular}{c|c|ccc}
\hline
\multirow{2}{*}{Model}  & \multirow{2}{*}{$\epsilon$}   & \multicolumn{3}{c}{Perplexity (Test Data)} \\ 
                        &                       & Base      & Compression       & Total   \\ \hline
HAQAE                   & -                     & -         & -                 & 21.38 $\pm$ 0.25   \\ \hdashline

SSDVAE                  & \multirow{2}{*}{0.9}  & -         & -                 & 19.84 $\pm$ 0.52   \\ 
\textit{grouping}                    &                       & \textbf{\textit{19.44 $\pm$ 0.5}}     & 31.36 $\pm$ 0.85             & 24.69 $\pm$ 0.64   \\ \hdashline
% ours+evt                &                      & 20.01    & 26.79             & 23.15   \\ \hdashline

SSDVAE                  & \multirow{2}{*}{0.7}  & -         & -                 & 21.19 $\pm$ 0.76   \\ 
\textit{grouping}                    &                       & \textit{20.13 $\pm$ 1.45}     & 29.7 $\pm$ 0.51             & 24.43 $\pm$ 0.84   \\ \hdashline
% ours+evt                &                      & 19.98    & 30.30             & 24.60   \\ \hdashline

SSDVAE                  & \multirow{2}{*}{0.5}  & -         & -                 & 31.11 $\pm$ 0.85   \\ 
\textit{grouping}                    &                       & \textit{21.52 $\pm$ 0.72}     & 31.62 $\pm$ 0.51             & 26.08 $\pm$ 0.39   \\ \hdashline
% ours+evt                &                      & 22.83    & 36.17             & 28.94   \\ \hdashline

SSDVAE                  & \multirow{2}{*}{0.4}  & -         & -                 & 33.12 $\pm$ 0.54   \\ 
\textit{grouping}                    &                       & \textit{23.42 $\pm$ 0.59}     & 30.16 $\pm$ 4.2             & 27.45 $\pm$ 0.66   \\ \hdashline
% ours+evt                &                      & 28.13    & 37.32             & 32.40   \\ \hdashline

SSDVAE                  & \multirow{2}{*}{0.2}  & -         & -                 & 33.31 $\pm$ 0.63   \\ 
\textit{grouping}                    &                       & \textit{28.17 $\pm$ 2.26}     & 34.17 $\pm$ 0.98             & 31 $\pm$ 1.31   \\ \hline
% ours+evt                &                      & 33.34    & 41.79             & 37.39   \\ \hline
\end{tabular}%
}
\caption{Per-word perplexity for test data (lower is better). For each observation probability ($\epsilon$), the best perplexity is in \textit{italic} form. The best of all of them is \textbf{bold} form. See \cref{sec:exp_3.13-app}}
\label{exp_3.13_ppl}
\end{table}

\begin{table}[t]
\resizebox{\columnwidth}{!}{%
\begin{tabular}{c|c|ccc}
\hline
\multirow{2}{*}{Model}  & \multirow{2}{*}{$\epsilon$}   & \multicolumn{3}{c}{Wikipedia INC (Test Data)} \\ 
                        &                       & Base     & Compression    & Total   \\ \hline
HAQAE                   & -                     & -        & -              & 24.88 $\pm$ 1.35   \\ \hdashline

SSDVAE                  & \multirow{2}{*}{0.9}  & -        & -              & 35.56 $\pm$ 1.70   \\ 
\textit{grouping}                    &                       & \textit{40.76 $\pm$ 2.86}    & 28.23 $\pm$ 1.04          & 39.4 $\pm$ 1.59   \\ \hdashline
% ours+evt              &                       & \textit{47.00}    & 38.75          & 45.80   \\ \hdashline

SSDVAE                  & \multirow{2}{*}{0.7}  & -        & -              & \textit{39.08 $\pm$ 1.55}   \\ 
\textit{grouping}                    &                       & 38.09 $\pm$ 5.6    & 26.55 $\pm$ 0.51          & 37.83 $\pm$ 5.08   \\ \hdashline
% ours+evt              &                       & 33.20    & 38.50          & 37.80   \\ \hdashline

SSDVAE                  & \multirow{2}{*}{0.5}  & -        & -              & \textit{40.18 $\pm$ 0.90}   \\ 
\textit{grouping}                    &                       & 39.5 $\pm$ 3.45    & 25.61 $\pm$ 0.96          & 37.86 $\pm$ 2.56   \\ \hdashline
% ours+evt              &                       & 34.90             & 40.65          & 40.55   \\ \hdashline

SSDVAE                  & \multirow{2}{*}{0.4}  & -                 & -              & \textit{47.88 $\pm$ 3.59}   \\ 
\textit{grouping}                    &                       & 43.83 $\pm$ 1.75             & 24.79 $\pm$ 0.43          & 42.16 $\pm$ 1.43   \\ \hdashline
% ours+evt              &                       & 41.40             & 41.75          & 45.00   \\ \hdashline

SSDVAE                  & \multirow{2}{*}{0.2}  & -                 & -              & 44.38 $\pm$ 2.10   \\ 
\textit{grouping}                    &                       & \textit{\textbf{48.88 $\pm$ 1.37}}    & 26.64 $\pm$ 0.98          & 46.81 $\pm$ 1.67   \\ \hline
% ours+evt              &                       & 46.40    & 38.25          & 45.90   \\ \hline
\end{tabular}%
}
\caption{Wikipedia Inverse Narrative Cloze Score for test data (higher is better). For each observation probability ($\epsilon$), the best score is in \textit{italic} form. The best of all of them is \textbf{bold} form. See \cref{sec:exp_3.13-app}}
\label{exp_3.13_winv}
\end{table}

Looking at the perplexity results (\cref{exp_3.13_ppl}) of this experiment, we can see that the base layer outperforms both baselines across observation lavels. Additionally, while we see the intuitive result that higher levels of frame observation during training improves perplexity, we see the largest relative improvments for $\epsilon=0.5$ and $\epsilon=0.4$. This suggests that our hierarchical model is able to effectively leverage the semantic ontology, even when $40\%$ of events do not have observed frames. %

%Also, the average score is better for lower observation probabilities ($0.5, 0.4, 0.2$). 

We see broadly similar patterns for inverse narrative cloze, with our approach outperforming both baselines. First, our performance is highest with the lowest observation level. Second, aside from when $90\%$ of the events have observed frames, as $\epsilon$ decreases, so does our model's variance, while the previous state-of-the-art's increases. Taken together, these results suggest that our model is better able to use the provided semantic ontology and make better longer range predictions, even with limited observations. %Second, result in most of the observation probability. 
Together with the perplexity improvements, these results reaffirm our assumption that the compression layer gives a subtle but strong signal that improves generative performance.

\section{Ablation Study}
\label{sec:ablation}

\begin{table*}[t]
\centering
\resizebox{\textwidth}{!}{%
\begin{tabular}{c|c|ccc|ccc}
\hline
\multirow{2}{*}{Model}  & \multirow{2}{*}{$\epsilon$}   & \multicolumn{3}{c}{Perplexity (Test Data)} &  \multicolumn{3}{c}{Wikipedia INC (Test Data)}\\ 
                        &                               & Base      & Compression    & Total                & Base      & Compression       & Total\\ \hline
HAQAE                   & -                             & -         & -              & 21.39 $\pm$ 0.25                & -         & -                 & 24.88 $\pm$ 1.35 \\ \hdashline

SSDVAE                  & \multirow{5}{*}{0.9}          & -         & -              & 19.84 $\pm$ 0.52                & -         & -                 & 35.56 $\pm$ 1.70 \\ 
$ours_{encdec}$         &                               & 26.25 $\pm$ 0.12     & 26.59 $\pm$ 0.13          & 26.42 $\pm$ 0.12                & 38.35 $\pm$ 1.66     & 38.42 $\pm$ 1.53             & 38.28 $\pm$ 1.65 \\ 
$ours_{frame}$          &                               & 20.94 $\pm$ 0.86     & 37.01 $\pm$ 1.55          & 27.83 $\pm$ 0.85                & 41.82 $\pm$ 2.44     & 28.37 $\pm$ 4.09             & 39.97 $\pm$ 1.16 \\
$ours_{sum}$            &                               & \textit{18.63 $\pm$ 0.24}     & 32.02 $\pm$ 4.46          & 24.38 $\pm$ 1.59                & 40.88 $\pm$ 0.25     & 36.15 $\pm$ 11.71             & 42.65 $\pm$ 5.02 \\ 
$ours_{cat}$            &                               & 19.34 $\pm$ 1.04     & 31.25 $\pm$ 2.07          & 24.54 $\pm$ 0.23                & \textit{44.05 $\pm$ 0.61}     & 25.43 $\pm$ 4.73             & 37.53 $\pm$ 4.54 \\ \hdashline

SSDVAE                  & \multirow{5}{*}{0.7}          & -         & -              & 21.19 $\pm$ 0.76                & -         & -                 & 39.08 $\pm$ 1.55 \\ 
$ours_{encdec}$         &                               & 27.15 $\pm$ 0.64     & 27.61 $\pm$ 0.64          & 27.38 $\pm$ 0.64                & 40.68 $\pm$ 1.78     & 40.37 $\pm$ 1.27             & 40.52 $\pm$ 1.43 \\ 
$ours_{frame}$          &                               & 20.77 $\pm$ 0.2     & 38.75 $\pm$ 1.18          & 28.37 $\pm$ 0.33                & 41.38 $\pm$ 3.48     & 33.22 $\pm$ 4.4             & 41.71 $\pm$ 2.75 \\ 
$ours_{sum}$            &                               & \textit{19.51 $\pm$ 0.5}     & 30.37 $\pm$ 3.29          & 24.33 $\pm$ 1.61                & 41.68 $\pm$ 1.25     & 31.77 $\pm$ 10.34             & 40.92 $\pm$ 5.04 \\ 
$ours_{cat}$            &                               & 20.17 $\pm$ 0.42     & 30.04 $\pm$ 2.89          & 24.59 $\pm$ 1.09                & \textit{43.42 $\pm$ 1.53}     & 28.15 $\pm$ 5.53             & 39.63 $\pm$ 3.45 \\ \hdashline

SSDVAE                  & \multirow{5}{*}{0.5}          & -         & -              & 31.11 $\pm$ 0.85                & -         & -                 & 40.18 $\pm$ 0.90 \\ 
$ours_{encdec}$         &                               & 26.54 $\pm$ 1.68     & 28.79 $\pm$ 1.55          & 27.65 $\pm$ 1.61                & 37.02 $\pm$ 5.75     & 37.03 $\pm$ 5.7             & 36.9 $\pm$ 5.9 \\ 
$ours_{frame}$          &                               & 19.55 $\pm$ 0.89     & 37.84 $\pm$ 1.72          & 27.19 $\pm$ 0.98                & \textit{45.48 $\pm$ 3.63}     & 27.9 $\pm$ 1.68             & 40.7 $\pm$ 3.55 \\ 
$ours_{sum}$            &                               & \textit{19.15 $\pm$ 0.38}     & 30.58 $\pm$ 1.28          & 24.19 $\pm$ 0.57                & 41.03 $\pm$ 1.32     & 43.37 $\pm$ 2.03             & \textit{46.83 $\pm$ 1.55} \\ 
$ours_{cat}$            &                               & 19.59 $\pm$ 0.22     & 30.39 $\pm$ 1.49          & 24.4 $\pm$ 0.6                & 41.45 $\pm$ 2.05     & 26.12 $\pm$ 4             & 38.57 $\pm$ 5.59 \\ \hdashline

SSDVAE                  & \multirow{5}{*}{0.4}          & -         & -              & 33.12 $\pm$ 0.54                & -         & -                 & \textit{\textbf{47.88 $\pm$ 3.59}} \\ 
$ours_{encdec}$         &                               & 25.56 $\pm$ 0.53     & 28.03 $\pm$ 0.47          & 26.77 $\pm$ 0.5                & 36.52 $\pm$ 3.06     & 36.23 $\pm$ 2.85             & 36.57 $\pm$ 2.97 \\ 
$ours_{frame}$          &                               & 19.6 $\pm$ 1.16     & 38.03 $\pm$ 0.74          & 27.29 $\pm$ 0.58                & 38.13 $\pm$ 2.55     & 26.78 $\pm$ 3.21             & 37.18 $\pm$ 0.73 \\ 
$ours_{sum}$            &                               & 18.79 $\pm$ 0.98     & 32.09 $\pm$ 1.27          & 24.56 $\pm$ 1.04                & 43.33 $\pm$ 0.88     & 37.47 $\pm$ 14.06             & 45.82 $\pm$ 4.8 \\ 
$ours_{cat}$            &                               & \textit{18.74 $\pm$ 0.83}     & 32.1 $\pm$ 2.14          & 24.52 $\pm$ 1.14                & 42.28 $\pm$ 3.73     & 32.37 $\pm$ 9.64             & 43.2 $\pm$ 2.66 \\ \hdashline

SSDVAE                  & \multirow{5}{*}{0.2}          & -         & -              & 33.31 $\pm$ 0.63                & -         & -                 & \textit{44.38 $\pm$ 2.10} \\ 
$ours_{encdec}$         &                               & 25.62 $\pm$ 0.31     & 30.85 $\pm$ 0.17          & 28.12 $\pm$ 0.1                & 38.1 $\pm$ 3.1     & 38.32 $\pm$ 3.37             & 38.27 $\pm$ 3.22 \\
$ours_{frame}$          &                               & 18.63 $\pm$ 0.75     & 38.68 $\pm$ 0.36          & 26.84 $\pm$ 0.65                & 41.45 $\pm$ 2.33     & 29.62 $\pm$ 0.98             & 40.43 $\pm$ 2.95 \\ 
$ours_{sum}$            &                               & \textbf{\textit{17.1 $\pm$ 0.21}}     & 29.19 $\pm$ 3.06          & 22.33 $\pm$ 1.31                & 39.25 $\pm$ 4.42     & 31.65 $\pm$ 11.88             & 40.45 $\pm$ 7.24 \\ 
$ours_{cat}$            &                               & 17.21 $\pm$ 0.65     & 29.6 $\pm$ 2.78          & 22.56 $\pm$ 1.18                & 38.55 $\pm$ 0.41     & 20.45 $\pm$ 9.19             & 29.77 $\pm$ 9.34 \\ \hline
\end{tabular}%
}
\caption{Wikipedia Inverse Narrative Cloze Score for test data (higher is better). For each observation probability ($\epsilon$), the best score is in \textit{italic} form. The best of all of them is \textbf{bold} form. See \cref{exp_3.18_1}, \cref{exp_3.18_2}, \cref{exp_3.18_3}}
\label{exp_3.18_all}
\end{table*}

% exp 3.18
\subsection{Impact of parameter sharing of encoder and decoder} \label{exp_3.18_1}
To find out the importance of multiple encoders and decoders on two layers, we have used shared parameters on both of them and see the effect on the result. The result for this experiment ($ours_{encdec}$) is reported on \cref{exp_3.18_all}. We can see a substantial drop in the result, especially on the INC score for low perplexity scores (0.5, 0.4, 0.2).

% exp 3.18
\subsection{Impact of parameter sharing of frame embedding} \label{exp_3.18_2}
To determine the importance of multiple frame embedding weights for each layer, we have used one shared frame embedding layer across both layers. We compute results across three seeds. The result for this experiment ($ours_{frame}$) is reported on \cref{exp_3.18_all}. Similar to the encoder-decoder, we can see a substantial decrease in the INC score.

% exp 3.18
\subsection{Impact of summation or concatenation of both layer encoding} \label{exp_3.18_3}
To illustrate if both layer encodings altogether can improve the result, we have done two experiments, one with the summation of both layers encodings ($ours_{sum}$) and another with only concatenation of both layer encodings ($ours_{cat}$). Both experiments' results are reported on \cref{exp_3.18_all}. Both of the models have a large drop on INC, which demonstrates the importance of the performance of the individual encoding.

\end{document}